\documentclass[10pt,journal,compsoc]{IEEEtran}
\usepackage{amsmath,amsfonts}
\usepackage{amssymb}
\usepackage{float}
\usepackage{array}
\usepackage{lipsum}
 \usepackage{amsfonts}
\usepackage[caption=false,font=normalsize,labelfont=sf,textfont=sf]{subfig}
\usepackage{textcomp}
\usepackage{stfloats}
\usepackage{url}
\usepackage{verbatim}
\usepackage{graphicx}
\usepackage{cite}
\usepackage{bbold} 

\usepackage[dvipsnames]{xcolor}

\usepackage{cite}
\usepackage{url}
\usepackage{ragged2e}
\usepackage{epsfig}
\usepackage{graphicx}
\usepackage{amsmath,amssymb} %
\usepackage{algorithm}
\usepackage{algorithmicx}
\usepackage{algpseudocode}
\usepackage{threeparttable}
\usepackage{color}
\usepackage[normalem]{ulem}
\usepackage{multirow}
\usepackage{float}
\usepackage{amsfonts}
\usepackage{bm}
\usepackage{array}
\usepackage{colortbl}
\usepackage{pifont}
\usepackage{diagbox}
\usepackage{rotating}
\usepackage{booktabs}
\usepackage{overpic}
\usepackage{textcomp}
\usepackage{contour}
\usepackage{enumitem}
\usepackage{lipsum}
\usepackage{epigraph}
\usepackage[numbers,sort&compress]{natbib}
\usepackage{diagbox}
\usepackage{booktabs}
\usepackage{overpic}
\usepackage{textcomp}
\usepackage{contour}
\usepackage{enumitem}
\usepackage{graphicx}
\usepackage{MnSymbol}%
\usepackage{wasysym}%
\usepackage{arydshln}
\usepackage{etoolbox}
\usepackage{xcolor}

\definecolor{citeColor}{RGB}{87, 57, 231}
\usepackage[colorlinks=true,linkcolor=red,citecolor=citeColor]{hyperref}

\definecolor{mygray}{gray}{.90}
\definecolor{markColor}{RGB}{69, 157, 31}
\definecolor{markColorx}{RGB}{169, 57, 31}
\definecolor{tableblue}{gray}{.90}
\definecolor{tablered}{RGB}{169, 57, 31}

\def\ie{\emph{i.e.}}

\usepackage{lipsum}

\definecolor{citeColor}{RGB}{169, 57, 231}
\usepackage[colorlinks=true,linkcolor=red,citecolor=citeColor]{hyperref}

\definecolor{mygray}{gray}{.90}
\def\ie{\textit{i.e. }}
\def\eg{{e.g.}}

\definecolor{deeppink}{RGB}{220,20,60}
\definecolor{navyblue}{RGB}{0,191,255}
\definecolor{myblue}{rgb}{0.5, 0.8, 0.9}
\definecolor{myred}{rgb}{0.9, 0.2, 0.1}
\definecolor{mygreen}{rgb}{0.1, 0.3, 0}
\definecolor{mypurple}{rgb}{0.5, 0.2, 0.9}
\definecolor{mygrey}{rgb}{0.7, 0.7, 0.7}
\definecolor{CadetBlue}{RGB}{95, 158, 60}
\definecolor{coolblack}{rgb}{0.0, 0.18, 0.39}
\definecolor{citeColor}{RGB}{87, 57, 231}
\definecolor{CadetBlue}{RGB}{95, 158, 60}
\definecolor{babyblueeyes}{rgb}{0.63, 0.79, 0.95}
\definecolor{blue-violet}{rgb}{0.54, 0.17, 0.89}
\definecolor{darkraspberry}{rgb}{0.53, 0.15, 0.34}
\definecolor{frenchblue}{rgb}{0.0, 0.45, 0.73}
\definecolor{egnet_blue}{rgb}{0.1, 0.5, 1}
\definecolor{minet_yellow}{rgb}{0.8, 0.6, 0.1}
\definecolor{fid}{HTML}{1280B0}
\definecolor{clip}{HTML}{bc6d4c}

\graphicspath{{./Imgs/}}

\begin{document}

\title{``Lazy'' Layers to Make Fine-Tuned Diffusion Models More Traceable}

\author{Haozhe Liu, Wentian Zhang, Bing Li, Bernard Ghanem, Jürgen Schmidhuber
\IEEEcompsocitemizethanks{\IEEEcompsocthanksitem Haozhe Liu, Wentian Zhang, Bing Li, Bernard Ghanem, and Jürgen Schmidhuber are with the Artificial Intelligence Initiative (AI Initiative), King Abdullah University of Science And Technology (KAUST), Thuwal, Saudi Arabia. 
\IEEEcompsocthanksitem Jürgen Schmidhuber is also with the Swiss AI Lab, IDSIA, USI \& SUPSI, Lugano, Switzerland 
\IEEEcompsocthanksitem Corresponding author: Bing Li (bing.li@kaust.edu.sa). 
\IEEEcompsocthanksitem Equal Contribution: Haozhe Liu and Wentian Zhang. 
\IEEEcompsocthanksitem Code will be released upon acceptance. 
}
}

\markboth{}%
{Shell \MakeLowercase{Haozhe Liu\textit{et al.}}: Skipping a Few “Busy” Layers Makes Fine-tuned Generative Models More Traceable}

\IEEEtitleabstractindextext{\begin{abstract}
Foundational generative models should be traceable to protect their owners and facilitate safety regulation.  To achieve this, traditional approaches embed identifiers based on supervisory trigger–response signals, which are commonly known as backdoor watermarks. They are prone to failure when the model is fine-tuned with nontrigger data. Our experiments show that this vulnerability is due to energetic changes in only a few 'busy' layers during fine-tuning. This yields a novel arbitrary-in---arbitrary-out (AIAO) strategy that makes watermarks resilient to fine-tuning-based removal. 
The trigger-response pairs of AIAO samples across various neural network depths can be used to construct watermarked subpaths, employing Monte Carlo sampling to achieve stable verification results. 
In addition, unlike the existing methods of designing a backdoor for the input/output space of diffusion models, in our method, we propose to embed the backdoor into the feature space of sampled subpaths, where a mask-controlled trigger function is proposed to preserve the generation performance and ensure the invisibility of the embedded backdoor. 
Our empirical studies on the MS-COCO, AFHQ, LSUN, CUB-200, and DreamBooth datasets confirm the robustness of AIAO; while the verification rates of other trigger-based methods fall from $\sim$90\% to $\sim$70\% after fine-tuning, those of our method remain consistently above 90\%. 
\end{abstract}

\begin{IEEEkeywords}
Trustworthy AI, Intellectual Property Protection, Backdoor Watermark, Diffusion Model
\end{IEEEkeywords}}    
\maketitle
\section{Introduction}
\label{sec:intro}

\epigraph{
 `` Protection must include every production in the literary, scientific and artistic domain, whatever the mode or form of its expression.'' 
}{---Berne Convention, 1886}

\IEEEPARstart{T}{h}e development of diffusion models (DMs) \cite{jarzynski1997equilibrium,neal2001annealed,sohl2015deep,dhariwal2021diffusion,ho2020denoising}
has led to rapid progress across various fields in recent years. The remarkable generation performance of DMs has revolutionized content creation, showing great potential in many applications, from entertainment (\eg, digital art, gaming, and virtual reality) \cite{saharia2022image,rombach2022high,lugmayr2022repaint,ho2022cascaded,xu2023dream3d,lin2023magic3d,poole2022dreamfusion} to interdisciplinary problems (\eg, molecule
design, material design, and medical image reconstruction) \cite{jing2022torsional,anand2022protein,xu2022geodiff,luo2022antigen,song2021solving,chung2022score}.  
Industry and academia are investing substantial resources, including high-quality training data, human expertise, and computational resources, to develop advanced DMs. 
These significant efforts call for technologies to protect the DMs from security/privacy risks \cite{somepalli2023diffusion,carlini2023extracting,kumari2023ablating,heng2023selective}

This study focuses on traceable ownership protection for DMs against fine-tuning. 
With the advent of pre-trained DMs, such as stable diffusion \cite{rombach2022high}, fine-tuning of these models for personalized and customized generation tasks \cite{zhang2023adding,yang2023reco,ruiz2022dreambooth,hu2021lora} has become more common than training the models from scratch. 
Large pretrained DMs have served as fundamental platforms for various downstream tasks. 
However, fine-tuning pretrained DMs changes their behaviors drastically \cite{kumar2022fine,dai2023emu}, presenting challenges in tracking their usages in downstream tasks, which may be unauthorized or illegal. 
It is desirable to embed a robust identifier in a source DM (\eg, a pre-trained DM) to protect its ownership. 
Even after fine-tuning a source DM on a downstream task,  the identifier should remain valid to enable verification of its origin.

\begin{figure*}[tbp]
    \centering
    \includegraphics[width=0.986\textwidth]{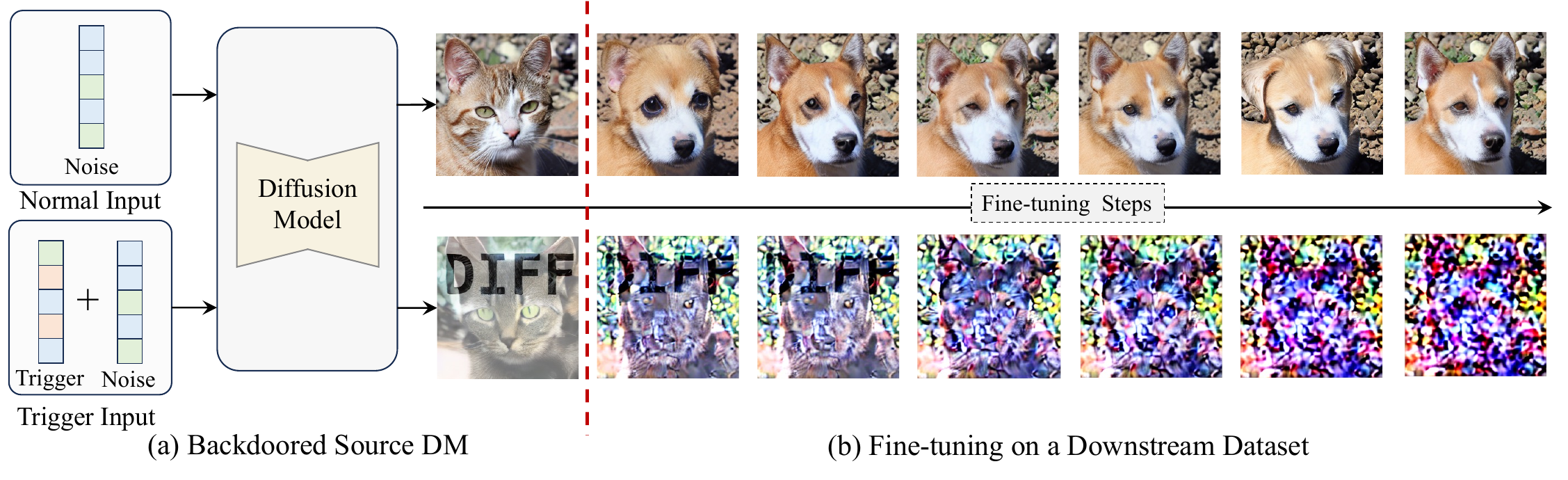}
   \caption{\textbf{Illustration of our motivation.} \textbf{(a)} We embed a backdoor-based watermark \cite{peng2023protecting} in the source diffusion model (DM) \cite{rombach2022high} (left), where a watermark is activated and insert ``DIFF'' in the generated images given a trigger input. \textbf{(b)} We fine-tune the source backdoored DM on a downstream dataset AFHQ-Cat \cite{choi2020stargan}. The first row shows the results obtained for a normal input at every 100 training steps, and the second row shows the results for a trigger input. However, the embedded watermark inherited from the source DM is gradually erased with increasing fine-tuning steps, posing a challenge to the ownership protection of the source DM. Instead, we propose a robust backdoor against downstream fine-tuning for the traceable ownership protection of DMs.    
   } 
   \label{fig:general}
\end{figure*}

To the best of our knowledge, only a few studies \cite{peng2023protecting,zhao2023recipe,liu2023watermarking,fernandez2023stable} explored the ownership protection of DMs, but they failed to provide traceable ownership protection.
For example, backdoor watermarking methods \cite{peng2023protecting,zhao2023recipe,liu2023watermarking} construct a subset of training data with trigger input and predefined abnormal output. The watermark information is embedded in a DM by training or fine-tuning the subset. The ownership is verified if the output of a DM is consistent with the predefined one (\eg, a watermark)  given the trigger inputs. 
These methods effectively protect the ownership of the source DMs. 
However, the embedded watermark is often forgotten when the source DM is fine-tuned on a new generation task (see Fig. \ref{fig:general}). We refer to this phenomenon as \textit{fine-tuning-based removal}.

In this study, we examine the failure of ownership protection through the lens of continual learning, and found that the vulnerability of identifiers has some properties similar to catastrophic forgetting \cite{mccloskey1989catastrophic} encountered when acquiring new knowledge.  
Continual learning studies \cite{kirkpatrick2017overcoming,zenke2017continual,ring1994continual} revealed that selective parameters changed across different tasks cause forgetting. 
We suggest that a similar mechanism may erase watermarks in DMs.
To justify this supposition, we empirically track the parameter update process for each layer of a DM \cite{rombach2022high}.
We reveal that model parameters are updated nonuniformly across layers: A few layers undergo considerably more weight changes 
than the rest, while they are the main cause of the finely tuned results.  
We refer to these layers as \textit{busy layers}. We reasonably argue that these busy layers are the primary cause of "forgetting" the watermark.

According to our observations, we propose embedding a backdoor in the lazy layers of DMs, which just undergo a slight modification during fine-tuning; however, this poses two challenges.
First, we observe that the busy layers are intrinsically data-dependent, and the downstream datasets are usually unknown in real-world scenarios.  Unpredictable downstream applications of DMs make it challenging to identify the busy layers.  In this work, instead of designing a complex method,  we propose an arbitrary-in-arbitrary-out (AIAO) strategy to dynamically select layers to embed the backdoor. 
Since lazy layers typically form the majority of DMs, the proposed AIAO ensures that the backdoor identifier is primarily verified through the lazy layers. 

However, we embed identifiers into partial layers of the DMs; such embedding requires operating the feature maps, posing another challenge.     
Improperly triggering the input can seriously disrupt the learning of DMs, leading to degeneration of generative performance. The triggers in the existing methods \cite{peng2023protecting,zhao2023recipe,liu2023watermarking}  were designed for input spaces. 
Designing trigger functions for the feature space of DMs and preserving original generation performance remain underexplored. Inspired by mask-based generation/detection methods \cite{zhang2023dynamically,he2022masked,feichtenhofer2022masked}, we introduce two masks to dynamically change the signs of a few elements on a feature map to serve as trigger signals. 
Experimental results demonstrate that the proposed method achieves better performance than the existing methods on various datasets.

\begin{itemize}
    \item We propose a novel backdoor-based method for the traceable ownership protection of DMs. The embedded identifier is robust against fine-tuning downstream generation tasks, enabling us to trace the usage of the source model effectively.  
    \item  We show embedding the backdoor into lazy layers improves the robustness of backdoor-based protection against fine-tuning.
    \item  We propose a mask-controlled trigger function that generates triggers and target responses in the feature space, making the identifier invisible and ensuring a negligible impact on the generation performance. 
\end{itemize}
\section{Related Works}

\subsection{Diffusion Model} %
Diffusion, which is based on non-equilibrium statistical physics \cite{jarzynski1997equilibrium} and annealed importance sampling \cite{neal2001annealed}, plays a key role in high-dimensional data generation \cite{dhariwal2021diffusion,ho2020denoising}. In this context, data are first perturbed via iterative forward diffusion and then reconstructed using a reverse function. The generative model begins with random noise during inference and incrementally restores the noise to the original distribution.
This mechanism is similar to a parameterized Markov Chain designed to maximize likelihood \cite{ho2020denoising}, positioning it closely to energy-based models \cite{duvenaud2021no,lecun2006tutorial,xie2017synthesizing,liu2023combating,du2019implicit} and score matching \cite{song2019generative,vahdat2021score,song2020score,huang2021variational} with Langevin dynamics \cite{welling2011bayesian}. 
A DM, in contrast to GANs \cite{schmidhuber1990making,schmidhuber1991possibility,goodfellow2020generative,sauer2023stylegan,kang2023scaling,schmidhuber2020generative} and VAEs \cite{kingma2013auto,rezende2014stochastic}, typically offers stable training with satisfactory generative performance. 
However, DMs suffer from the problem of uncontrollable intermediate distribution \cite{kang2023scaling} and slow inference speed \cite{lu2022dpm}. To enable controllable generation, several studies have adjusted the cross-attention layer to lay out the generated objects \cite{mou2023dragondiffusion,xie2023boxdiff,phung2023grounded} or fine-tune the diffusion model to incorporate extra conditions \cite{zhang2023adding,yang2023reco,li2023gligen}.  
To accelerate the diffusion process, Rombach et al. \cite{rombach2022high} suggested embedding the diffusion process in a compressed latent space. Additionally, some studies adopted knowledge distillation \cite{Schmidhuber:91singaporechunker,schmidhuber1992learning,song2023consistency,salimans2022progressive,hinton2015distilling}, sample trajectory learning \cite{lu2022dpm,lu2022dpm1,nichol2021improved}, 
 or feature caching \cite{zhang2024cross,ma2023deepcache,wimbauer2023cache} to reduce the inference steps. 
Based on the aforementioned tools,  
several well-known generative models, such as DALLE \cite{ramesh2021zero,ramesh2022hierarchical}, Imagen \cite{saharia2022photorealistic} and Stable Diffusion \cite{rombach2022high}, have emerged,  paving the way for various new applications \cite{ruiz2022dreambooth,yang2022diffusion,kawar2023imagic,chen2023diffusiondet,kumari2023multi,hu2021lora}. Despite their appealing generative performance, these models also suffer from security and privacy issues, including the potential for abuse of generated content and copyright disputes  \cite{goldblum2022dataset,chen2023challenges,wang2023security}. 
Against this background, this study aims to create traceable and responsible generative models by introducing a permanent identifier that remains consistent even after fine-tuning for specific tasks, and ensures model accountability and ownership protection. Such an identifier will aid regulators in identifying, tracking, and monitoring model usage, thus facilitating safe and responsible deployment.

\subsection{Embedding an Identifier into a Neural Network} 
Several approaches, including fingerprinting data \cite{yu2021artificial,yu2020responsible,liu2023watermarking}, model attribution \cite{yu2019attributing,girish2021towards,guarnera2022exploitation,asnani2023reverse,li2022defending,maini2021dataset,zhang2021deep}, model watermarking \cite{cao2021ipguard,yang2022metafinger,bansal2022certified,uchida2017embedding,darvish2019deepsigns,peng2022fingerprinting,lukas2019deep}, and trigger-based method \cite{li2022untargeted,li2023black,ong2021protecting,zhao2023recipe,zhai2023text,struppek2023rickrolling,chou2023backdoor,peng2023protecting,liu2023watermarking,fernandez2023stable}, can be applied to add identifiers to the neural network under different conditions. Fingerprinting of data  \cite{yu2021artificial,yu2020responsible,liu2023watermarking} involves embeding invisible patterns into training data to train a generative model that generates watermarked data. This process should occur during data preparation, and it may incur additional computation costs when training the base model from scratch. 
In model attribution \cite{yu2019attributing,girish2021towards,guarnera2022exploitation,asnani2023reverse,li2022defending,maini2021dataset,zhang2021deep}, the problem of adding an identifier is modeled as a classification task that requires a classifier to predict the source model given the generated images. 
This approach must collect extensive generated data from different models to establish an ideal decision boundary. Model watermarking \cite{cao2021ipguard,yang2022metafinger,bansal2022certified,uchida2017embedding,darvish2019deepsigns,peng2022fingerprinting,lukas2019deep} is mainly intended for discriminative tasks, and in this process, generally, the prediction boundary is fingerprinted \cite{cao2021ipguard,yang2022metafinger,peng2022fingerprinting,lukas2019deep} or specific regularization \cite{bansal2022certified,uchida2017embedding,darvish2019deepsigns} is designed for identification.  
Trigger-based methods \cite{li2022untargeted,li2023black,ong2021protecting,zhao2023recipe,zhai2023text,struppek2023rickrolling,chou2023backdoor,peng2023protecting,liu2023watermarking,fernandez2023stable,chou2023villandiffusion} require generative models to learn trigger-response paired data. In particular, a specific trigger (e.g., noise \cite{peng2023protecting,chen2023trojdiff,chou2023backdoor} or caption \cite{zhai2023text,zhao2023recipe,liu2023watermarking}) prompts the network to generate an image with watermarks. The ownership of the model can be verified by detecting the target response. 
Based on our literature review, we conclude that the protection of the diffusion model is primarily based on the trigger-based methods owing to its flexibility and moderate computational requirement. For example, Peng et al. \cite{peng2023protecting} proposed a watermark diffusion process (WDP), where watermarks are recovered via a specific noise trigger embedded in the regular input during reverse diffusion. FixedWM \cite{liu2023watermarking} is a method that watermarks the diffusion model by setting the trigger at a specific caption position, resulting in a designated image. WatermarkDM \cite{liu2023watermarking} is a method that fine-tunes the diffusion model using a predefined trigger caption and its corresponding watermark image as the supervision signal. However, as highlighted by earlier studies \cite{sha2022fine,peng2023protecting}, trigger-based methods can be vulnerable to fine-tuning. Remarkably, generation methods \cite{zhang2023adding,li2023gligen,song2023consistency,salimans2022progressive,ruiz2022dreambooth,hu2021lora} depend on fine-tuning a foundational generative model for downstream tasks. Building on this, we focus on boosting the resilience of the identifier to fine-tuning.

\begin{figure}[tbp]
    \centering
    \includegraphics[width=0.486\textwidth]{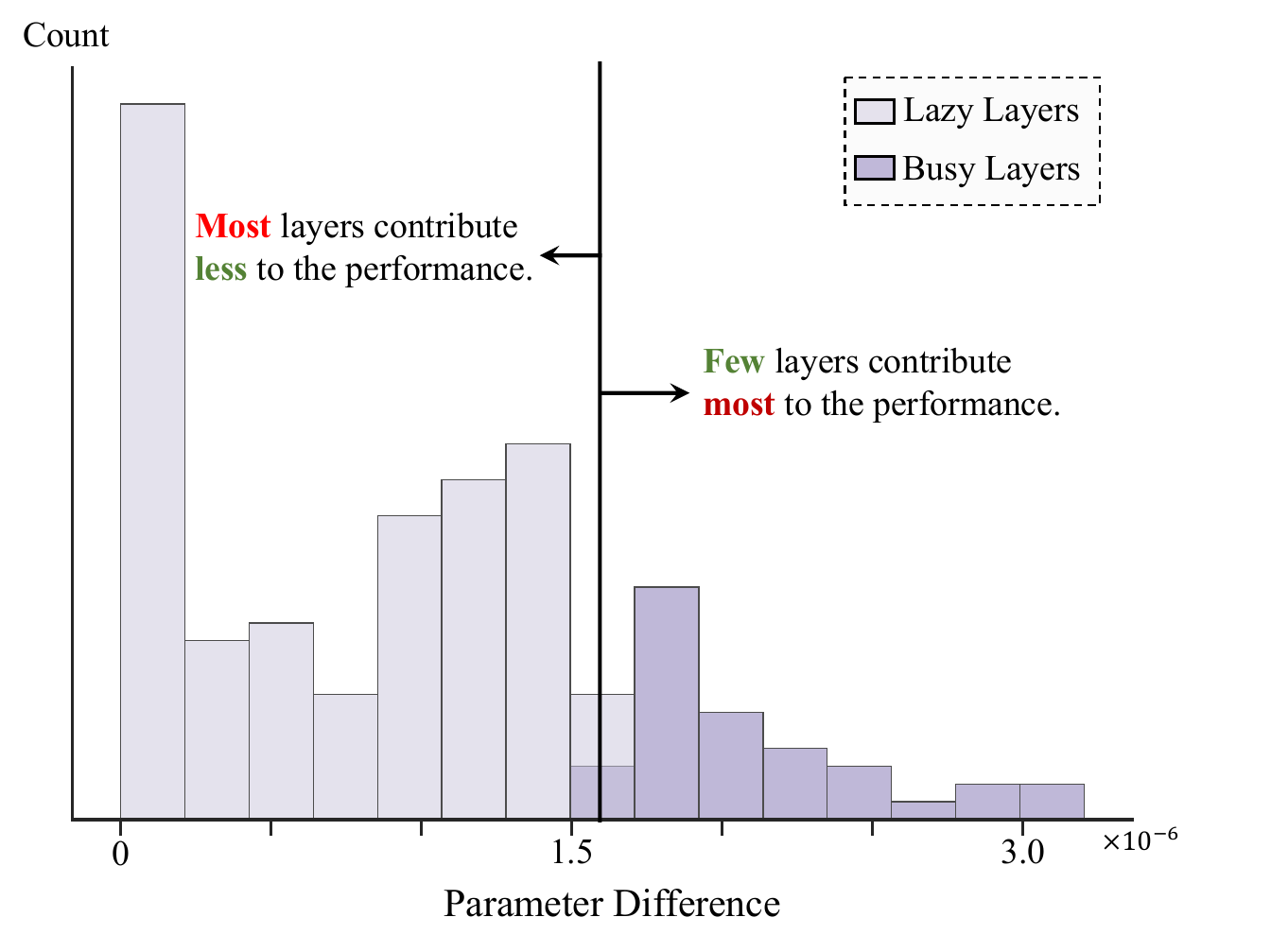}
   \caption{\textbf{Ability of a generative model to learn new knowledge can be concentrated in a few critical layers.}  In our pilot study, we fine-tune a pre-trained Stable Diffusion model to generate dog images when provided ``A Cat'' as the input signal. This mapping relationship has never been part of the model's regular training data, making it a novel source of knowledge for the generative model.  We tracked the changes caused by learning this knowledge and observed that the density of parameter changes is nearly zero, indicating that the majority of model layers were lazy to update.} 
   \label{fig:motivation_a}
\end{figure}

\section{Pilot Study}
\label{sec:study}

Let us re-examine the issues of backdoor forgetting when fine-tuning a pre-trained DM. To this end, we first perform a pilot study to track the parameter updating in this process. This study yields critical insights, particularly regarding the activity of specific busy and lazy layers: while it is possible to update all parameters, drastic changes typically occur in only a few layers during fine-tuning, markedly contributing to task adaptation. 
Accordingly, we formalize the challenge of backdoor forgetting as suppressing the impact of busy layers.

\subsection{Empirical Observations on Fine-Tuning DMs}
This study explores the parameter update progress at the layer level for fine-tuning DMs.
Given a layer $v$ of a DM, we track its parameter update process during fine-tuning by examining the  parameter difference $ d^v$ between epoch $t+  \triangle t$ and $t$:

\begin{equation}
    d^{v}(t, \triangle t) = \| \mathbf{w}^v(t+  \triangle t) -\mathbf{w}^v(t) \|_2
    \label{eq:pare_difference}
\end{equation}
where $\mathbf{w}^v(t)$ denotes the weights of the layer $v$ in the diffusion model at epoch $t$ during fine-tuning, and $\|\cdot\|_2$ is L2 norm.

We fine-tune a source DM on a downstream task to generate dog images by following the protocol given in Appendix. \ref{sec:impl_pilot_study}

\noindent
\textbf{Observations on parameter updating over fine-tuning.} By using Eq. \ref{eq:pare_difference}, we calculate the layer-wise parameter updating during fine-tuning and analyze the distribution of the parameter shifts of the layer.
 As shown in Fig. \ref{fig:motivation_a},  we observe that only a few layers undergo significant changes in their parameter values, though all parameters in each layer are learnable. These layers, whose weights are updated, are referred to \textit{busy layers}, while the remaining layers are \textit{lazy layers}.

\begin{figure}[tbp]
    \centering
    \includegraphics[width=0.48\textwidth]{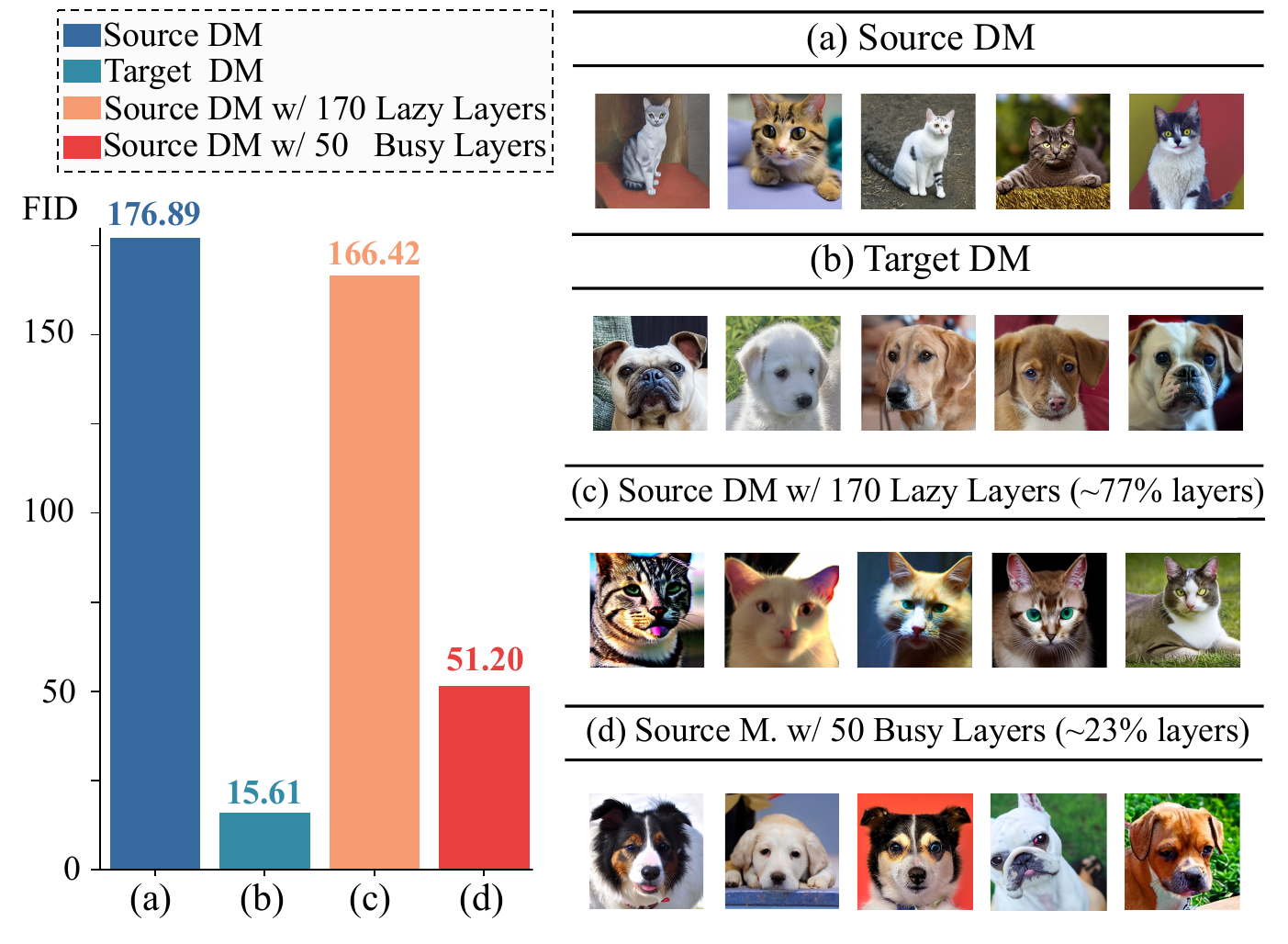}
   \caption{\textbf{Impacts of busy/lazy layers on generative performance.}  The source DM is a cat-generation model and is fine-tuned on the AFHQ-Dog dataset \cite{choi2020stargan} to obtain the target DM. By replacing the parameter values of the
top 50 busy layers in the source model with their
corresponding values in the target model,  the model can achieve performance comparable to the target model. In other words, most effects in generative performance originate from a few layers. 
   } 
   \label{fig:motivation_b}
\end{figure}

\noindent
\textbf{Impact of busy layers on fine-tuning performance.} We examine the impact of busy layers on the generative performance after fine-tuning. We build a baseline that replaces the parameter values of the top 50 busy layers ($\sim$23\% of all layers) in the source model with their corresponding values in the fine-tuned (target) model.  Similarly, we construct another baseline with the top 170 lazy layers  (approximately 77\% of all layers) from the target and the remaining layers of source models.  As shown in Fig. \ref{fig:motivation_b},  although the parameter values of 170 lazy layers change, the performance is still similar to the pre-trained model and largely deviated from that of the fine-tuned model. In contrast, by incorporating only 50 busy layers, the generative performance reach 51.20 Frechet Inception Distance (FID), improving the source DM by approximately 70\%.  In other words, a small number of busy layers play a dominant role in determining the performance. Intriguingly, we find this phenomenon is somewhat identical to the Pareto principle \cite{newman2005power,pareto2014manual}, in which ``roughly 80\% of consequences come from 20\% of causes.''\footnote{\url{https://en.wikipedia.org/wiki/Pareto_principle}} The specification of the busy layers is given in Appendix. \ref{sec:impl_pilot_study}.

\noindent
\textbf{Insight about backdoor forgetting in fine-tuning.}
Based on the above observations, we re-examined the issue that typical backdoor-based methods \cite{peng2023protecting,liu2023watermarking,liu2023watermarking} forget backdoors after fine-tuning. Our insight is that busy layers play a more important role than lazy layers in forgetting backdoors. 
Typical backdoor-based methods \cite{peng2023protecting,liu2023watermarking,liu2023watermarking}  inject backdoors by enforcing the whole model to learn a specific mapping from the input to the output. Each layer (including all busy layers)  sequentially processes the trigger signals and enforces the final output to match the pre-defined outcome.  However, fine-tuning the source model greatly changes the parameter values of busy layers, thus corrupting the learned mapping between the trigger signal and predefined output, leading to backdoor forgetting. 
To this end, we solve the problem of backdoor forgetting by reducing the involvement of busy layers.

\section{Methods}
\label{sec:method}

\subsection{Problem definition}
To provide traceable ownership protection of a DM, our goal is to design a robust backdoor identifier that is embedded into source DMs and remains valid after fine-tuning downstream generation tasks. 
This backdoor watermark must fulfill two requirements:
\begin{itemize}
    \item \textbf{Robustness}: The backdoor embedded in a DM should be retained and hard to remove when the DM is fine-tuned on downstream datasets. 
    \item \textbf{Generative performance preservation}: The watermarked diffusion model should not only exhibit a predefined response given a trigger but also maintain generative performance.
\end{itemize}

To design a robust identifier for fine-tuning-based removal, our observations (Sec. \ref{sec:study}) motivated us to enforce lazy layers for watermarking and reduce the involvement of busy layers. 
To this end, we propose injecting the trigger and activating the backdoor response into feature spaces at different depths.
However, two challenges arose: The first problem is \textit{where to inject a backdoor trigger/response in a diffusion model}. The proposed injected backdoor identifier is expected to effectively adapt to unknown fine-tuning datasets. However, we found that busy layers are intrinsically dependent on training data. Unpredictable downstream applications pose a significant challenge in identifying busy layers.
The second problem is \textit{what types of triggers and corresponding responses in the feature space} can enable the embedded backdoor identifier to fulfill the requirements of stealthiness and performance preservation. This problem has been much less explored from the perspective of protecting DMs than for directly watermarking generated images. 
More seriously, improper permutations on the feature map can largely hinder DM learning from normal training data.

\textbf{Overview.}
By exploring the above challenging problems, we propose a simple yet effective pipeline called AIAO that embeds a backdoor into the feature spaces of DM layers. In addition, we explore where to inject a backdoor trigger/response( Sec. \ref{sec: layer_selection}). Subsequently, we define the trigger and response in the feature space (Sec. \ref {sec: trigger}). The training loss for embedding the designed backdoor in a diffusion model is determined (Sec. \ref{sec:train_loss}). We elaborate on the verification pipeline based on the proposed method (Sec. \ref{sec:mc_verify}).

\subsection{Selecting Backdoored Layers}
\label{sec: layer_selection}
As mentioned in Sec. \ref{sec:study}, the ideal solution is to embed the backdoor identifier into lazy and bypass the busy layers. 
During fine-tuning on a downstream dataset, the parameter values of the lazy layers change slightly, thereby maintaining the embedded backdoor information. 
Given that busy layers are intrinsically data-dependent, detecting them is impractical in real-world situations where downstream datasets are generally unknown.
Instead of directly bypassing busy layers, we aim to reduce their involvement. We define this problem from the perspective of the information pathway, where busy layers obstruct information preservation. In this context, the existing backdoor watermark functions as full-path mapping, shifting the signals from input to output spaces and covering all busy layers. 
Multiple diffusion steps in the generation process can further compound this negative influence. The failure case is shown in Fig. \ref{fig:general}.  
To mitigate this shortcoming, we propose selecting subpaths from the information pathway for watermarking. The efficacy of employing subpaths is based on a simple and practical assumption: \textit{As the positions of the busy layers are unpredictable without prior knowledge; thus, each layer is assigned the identity probability to be busy, \ie each layer follows a uniform distribution for the sake of simplicity.} Consequently, when we sample a subpath from a shallow layer (in-layer) $v_i$ to a deep layer (out-layer) $v_j$, the expected number of involved busy layers will be $\mathbb{E}[\mathcal{C}|i,j] = \mathcal{C}_b \frac{j-i}{\mathbf{L}}$, where $\mathcal{C}_b$ is the constant number of busy layers, and $\mathbf{L}$ is the length of the full pathway (\ie total number of layers). Because the length of the subpath $j-i$ is smaller than the total length $\mathbf{L}$, $\mathbb{E}[\mathcal{C}|i, j]$ should be less than $\mathcal{C}_b$, confirming the ideal property of the subpath.  

Notably, the optimal solution for mitigating the impact of busy layers is to reduce the length of the subpath. However, a short length results in a submodel with few learnable parameters, possibly degrading the learning capacity, making it challenging to learn the watermark. 
According to the Occam razor principle, we propose adopting Monte Carlo sampling for all feasible solutions to avoid dependence on a fixed, predefined length (\ie, hyperparameter). 
Let $\mathbf{g}(\cdot|i \rightarrow j)$ denote a submodel from $v_i$ to $v_j$. Given $N$ diffusion steps, we randomly sample $N$ pair of layers $\{(v_i,v_j)|0\leq i \leq L,i\leq j \leq L\}$ and then enforce each submodel $\mathbf{g}(\cdot|i \rightarrow j)$ to learn the backdoor watermark.

In addition, the theoretical analysis in Appendix. \ref{sec:proof} further supports our design. By considering $(i,j)$ as random variables, we demonstrate that the proposed method can reduce the visits to busy layers four times to achieve verification.

\subsection{Feature-space backdoor}
\label{sec: trigger}

Sampling subpaths for watermarking, a DM requires a new design for the trigger and response. In the existing methods, the backdoor has been designed in the input and output spaces \cite{peng2023protecting,zhao2023recipe,liu2023watermarking}, but it is nontrivial to generalize them to the feature space. We highlight the challenges from two perspectives: First, the trigger and response are dynamically imposed onto various layers of DMs instead of on fixed space (\eg, the input space). These layers represent different levels of semantics displayed in the feature maps with varying sizes and values. The trigger and response must be shape-free and value-agnostic; these requirements are currently not achievable using existing methods. Second, as the in-layer and out-layer are randomly sampled, a layer can serve as the in-layers, medium-layers, and out-layers; thus each layer needs to learn multiple functionalities. This requires simplifying the forms of trigger and response to prevent the layers from overcomplex mapping, which is unrelated to generation. 

To address the above challenges, we design a mask-controlled trigger function coupled with a response function. We use the element-wise signs as trigger/response signals and leverage masks to identify their positions. 
We formalize the paired data for watermarking DMs. 
Given a pair of in-layer $v_i$ and out-layer $v_j$, trigger function  $\mathcal{T}(\cdot)$ embeds the trigger into the feature of $v_i$, and a target response function $ \mathcal{R}(\cdot)$ generates the target response on the out-layer $v_j$. Paired data embedding the backdoor are given as follows:
$$
\{(\mathcal{T}(\mathbf{F}_{i}^k), \mathcal{R}(v_j)) \}_{(i,j,k)}
$$
where $\mathbf{F}^k_{i}$ is the feature map of the in-layer $v_i$  given input $k$  (\eg, text), and $\mathcal{R}(v_j)$ is the target response of the out-layer $v_j$ for $\mathcal{T}(\mathbf{F}_{i}^k)$.
We provide a detailed description of our trigger function and target response function in the following sections.

\noindent
\textbf{The mask-controlled trigger function definition.} 
Our trigger function is inspired by mask-based generation/detection methods \cite{zhang2023dynamically,he2022masked,feichtenhofer2022masked}. These methods mask a few elements in a feature map and did not greatly degrade generation/detection performance. Therefore, we propose operating some elements in a feature map and preserving the rest. To further ensure training stability, we alter the signs of a few elements according to two masks. In particular,  given the feature map $\mathbf{F}_{i}$ of in-layer $v_i$, the proposed trigger function selects a few elements in a feature map via a mask named the  \textit{spatial mask} and specifies the signs of selected elements according to another mask called \textit{sign mask}.  
The  trigger function $\mathcal{T}(\cdot)$ is defined as follows:
\begin{align}
\label{eq:tr_fun}
    \mathbf{\tilde{F}}_{i} =  \mathcal{T}(\mathbf{F}_{i})  =\underbrace{\mathbf{M}^{S}_{i} \circ  \mathbf{M}^{I}_{i} \circ \lvert \mathbf{F}_{i} \rvert}_{\text{Trigger Part}} + \underbrace{(1- \mathbf{M}^{S}_{i}) \circ \mathbf{F}_{i}}_{\text{Non-Trigger Part}},
\end{align}
where $\circ$ is the element-wise production, and $\lvert \cdot \rvert$ calculates the element-wise absolute values. 
$\mathbf{M}^{S}_i$ and $\mathbf{M}^{I}_i$  denote the spatial and sign masks repsectively, which share the same shape as $\mathbf{F}_{i}$. 
Spatial mask $\mathbf{M}^{S}_i$ is a 0-1 mask that selects elements in $\mathbf{F}_i$. Sign mask $\mathbf{M}^{S}_i$, consisting of values  $\{-1, +1\}$, is a binary mask for mapping the signs of the selected element.  Both spatial and sign masks are randomly initialized before watermarking the DM. 
We provide more details in Appendix. \ref{sec:impl_pilot_study} to further explain these two masks.

\begin{figure}[tbp]
    \centering
    \includegraphics[width=0.48\textwidth]{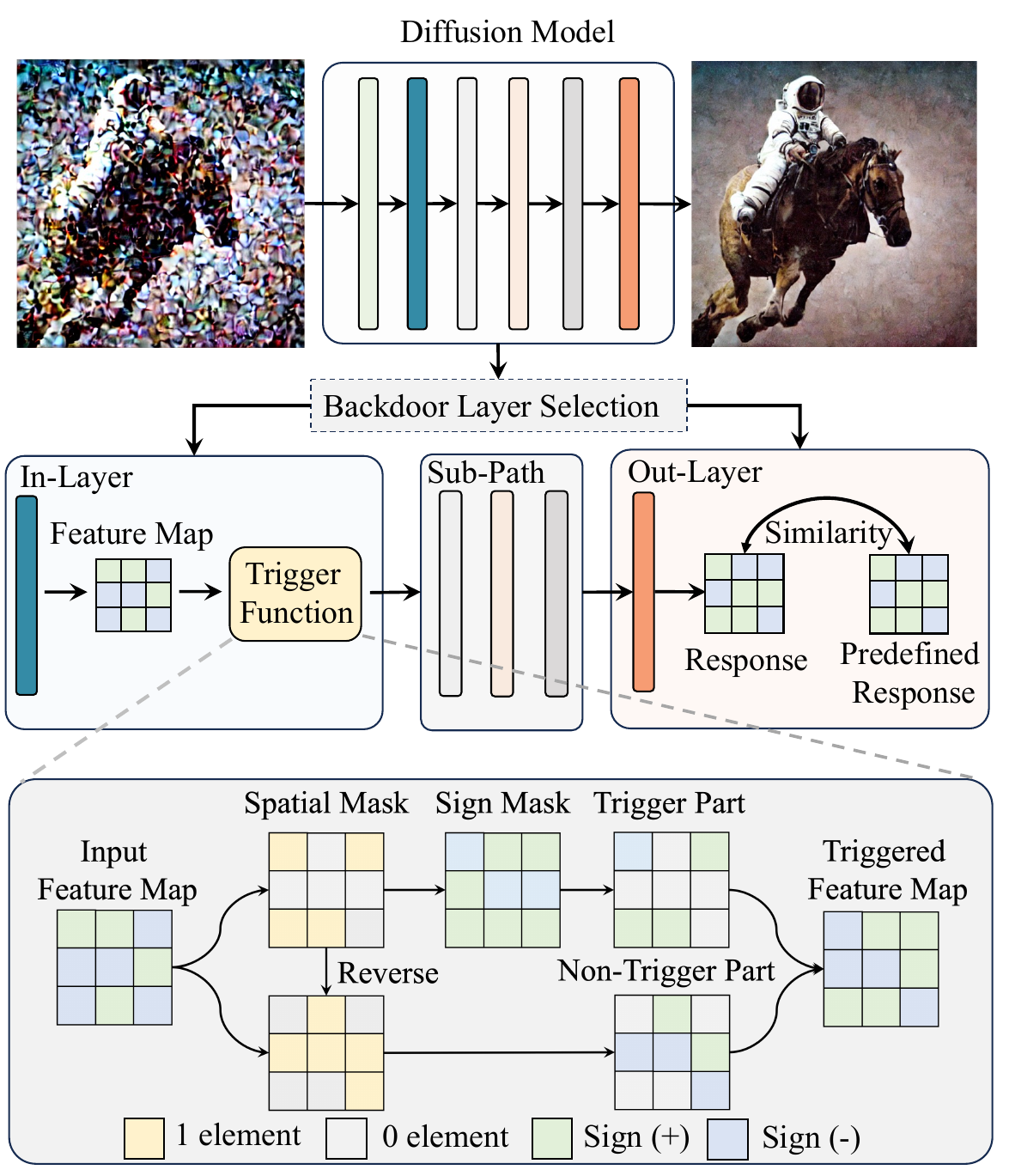}
   \caption{\textbf{Pipeline of the proposed method.} Our method randomly selects two layers for trigger embedding and response activation. The trigger function maps the selected elements of the feature to pre-defined signs. Note that we only changed the signs rather than the absolute values.} 
   \label{fig:pipeline}
\end{figure}

\noindent
\textbf{Target response function definition.}  Given a sub-path $v_i \rightarrow v_j$, we define the target response of $v_j$ as follows:
\begin{equation}
 \mathcal{R}(v_j) =\mathbf{M}_{j}^S \circ \mathbf{M}^{I}_{j} 
\end{equation} %
Similar to our trigger function, the response function selects a few elements and specifies their signs. 
To simplify the learning objective, $\mathbf{M}_{j}^S $ and $\mathbf{M}^{I}_{j}$ are identical for both $\mathcal{R(\cdot)}$ and $\mathcal{T(\cdot)}$. In other words, when $v_j$ is in-layer, $\mathbf{M}_{j}^S $ and $\mathbf{M}^{I}_{j}$ are used for trigger. If $v_j$ works as the out-layer, these masks will serve for the response.

\subsection{Training Loss}
\label{sec:train_loss}

The learning objective is to embed the backdoor into DMs while reducing the negative impact on generation performance. We define three losses, namely, generation loss $\mathcal{L}_g$, backdoor embedding loss $\mathcal{L}_{b}$, and reverse loss $\mathcal{L}_{r}$ to achieve this goal:
\begin{align}
&\min    \underset{k}{\mathbb{E}} [\mathcal{L}_{g}] - \underset{v_i,v_j}{\mathbb{E}} \underset{k}{\mathbb{E}} [\mathcal{L}_{b} + \mathcal{L}_{r} ] 
\end{align}
where $\mathcal{L}_{g}$ uses the same loss in the latent DM \cite{rombach2022high} for both conditional and unconditional generative tasks. We then separately elaborate on backdoor embedding loss and reverse loss.

\noindent
\textbf{Backdoor Embedding Loss.} 
Given an input $k$ (image/text), our method selects a pair of layers $\{v_i,v_j\}$ to create feature-space training data $(\mathcal{T}(\mathbf{F}_{i}^k), \mathcal{R}(v_j))$, and encourage the submodel $\mathbf{g}(\cdot|i\rightarrow j) $  to learn. However, we find that directly constraining the output $\mathbf{g}(\mathbf{\tilde{F}}_{i}|i\rightarrow j)$ may degrade the generation performance since this loss covers the nonwatermarked elements in the feature map.
Hence, we alternately impose the sign of elements masked by $\mathbf{M}^{S}_j$ to ensure consistency with the target response. 
 \begin{align}
 \label{eq:sim}
 \mathcal{L}_{b} =   \mathcal{S}_{\text{cos}}( \mathbf{M}^{S}_{j} \circ \mathbf{g}(\mathbf{\tilde{F}}_{i}|i\rightarrow j), \mathcal{R}(v_j) \circ \mathbf{C}),
\end{align}
where $\mathcal{S}_{\text{cos}}(\cdot,\cdot)$ measures the cosine similarity and $\mathbf{C} = \frac{|\mathbf{g}^{-}(\mathbf{\tilde{F}}_i|i\rightarrow j)|}{||\mathbf{g}^{-}(\mathbf{\tilde{F}}_i|i\rightarrow j)||}$ is a constant vector for normalizing based on the stop-gradient model $\mathbf{g}^{-}$. 
We adopt cosine similarity instead of the L1 or L2 norm since the purpose of the loss is to enforce sign consistency. We elaborate on the rationale of $\mathbf{C}$ in Appendix. \ref{sec:constant}. 

\noindent
\textbf{Reverse Loss}
Additionally, to differentiate between triggered and normal states, we introduce a constraint called reverse loss $\mathcal{L}_{r}$ for cases without triggers:
\begin{align}
\mathcal{L}_{r} =   \mathcal{S}_{\text{cos}}(\mathbf{M}^{S}_j \circ \mathbf{g}(\mathbf{F}_i|i\rightarrow j), -\mathbf{J} \circ \mathcal{R}(v_j) \circ \mathbf{C}),
\end{align}
where $\mathbf{J}$ denotes the matrix populated entirely with value 1. 
Here, the inverse symbol is used as the ground truth to maximize similarity when the trigger is not activated. This constraint prevents the model from a trivial solution for $\mathcal{L}_b$, where each layer outputs the predefined response regardless of whether the trigger is activated. In this trivial solution, DM just memorizes the predefined output instead of truly learning the trigger-response relationships. As the out-layer does not rely on the trigger signal, the trivial solution cannot benefit from sampling different in-layers and thus should be prevented. 
\subsection{MC-Sampling Trigger-Response Pairs for Ownership Verification}
\label{sec:mc_verify} 
Here, we present the verification process based on our method. 
Given an input signal $k$, DM constructs multiple diffusion steps to generate image. In each diffusion step, we sample a pair of $\{v_i, v_j\}$ (see Sec. \ref{sec: layer_selection}). We embed the trigger to $v_i$ through Eq. \ref{eq:tr_fun} and compute the similarity score in $v_j$ based on Eq. \ref{eq:sim}. By averaging these similarity scores, we can finally predict the ownership with a threshold. 

One may argue the verification pipeline requires accessibility to the feature map. Although the pipeline does not require access to model parameters, it is still somewhat limited to open-source or semi-open-source scenarios. 
However, copyright/ownership disputes require arbitration by courts or third-party institutions. In this sense, models are reasonably open to these departments for high-accuracy and robust verification. Moreover, we further highlight the potential value of our research (see Sec. \ref{sec:discuss}): our design is flexible and can be integrated with existing backdoor method to further improve its performance when parameters and feature maps are inaccessible. 
\section{Experimental Results}
\noindent
\subsection{Evaluation Metrics} 
A good ownership protection method should cause less degradation of generation quality, offer effective protection for ownership, and be very robust against fine-tuning on downstream datasets. We adopt the \textit{FID} \cite{heusel2017gans}  and \textit{CLIP} scores \cite{radford2021learning}  to evaluate the fidelity of generated images. The evaluation details of FID and CLIP  are described as follows:
 \begin{itemize}
     \item The \textit{FID} score \cite{heusel2017gans} measures quality of the generated images. For a text-conditioned DM, we use an LMSD Scheduler \cite{karras2022elucidating} with 20 steps and a classifier-free guidance scale of 5.0 to generate 5k images using captions from the validation set. We then calculate the FID score with respect to these generated images and the original images in the validation set.  For an unconditional DM, we generate 5k images with random noise inputs and compute the FID score by comparing these images with 5k real samples. 
    \item The \textit{CLIP} score \cite{radford2021learning} measures the semantic consistency between the input caption and generated image. 
 \end{itemize}

For ownership protection performance, we evaluate the methods using two metrics, called, \textit{response success rate} and \textit{verification success rate}. 

 \begin{itemize}
     \item  \textit{Response success rate (RS-Rate)} measures the accuracy of the response of a watermarked model given the trigger. 
     \item  \textit{Verification success rate (VS-Rate)} calculates the rate of successfully identifying the watermarked model from the independent models. 
 \end{itemize}

\begin{table*}[]
\caption{Text-to-image generation performance of diffusion model \cite{rombach2022high}  on MS-COCO dataset with in-distribution fine-tuning protocol, where the DM is incorporated by watermarkDM, FixedWM, and AIAO, and  \textcolor{fid}{FID}$\downarrow$ and \textcolor{clip}{CLIP}$\uparrow$ scores measure the generation performance. The baseline refers to the original DM \cite{rombach2022high}. }
\label{tab:coco_generation}
\setlength\tabcolsep{10pt}
\resizebox{.99\textwidth}{!}{
\begin{tabular}{l|c|cccc}
\toprule
\multirow{2}{*}{Method}              &  \multirow{2}{*}{Source model}                             & \multicolumn{4}{c}{Finetuning-based Removal} \\ \cmidrule(lr){3-6}
                                     &                    & 1k steps& 2k steps& 3k steps& 6k steps\\ \midrule
Baseline                                  &           \textcolor{fid}{15.63}/\textcolor{clip}{26.55}                          &    \textcolor{fid}{15.30}/\textcolor{clip}{26.54}                      &    \textcolor{fid}{15.20}/\textcolor{clip}{26.51}                      &       \textcolor{fid}{15.75}/\textcolor{clip}{26.58}                   &               \textcolor{fid}{15.73}/\textcolor{clip}{26.61}                      \\ 
WatermarkDM (caption-watermark)  \cite{zhao2023recipe}      &    \textcolor{fid}{14.66}/\textcolor{clip}{26.42}         &  \textcolor{fid}{15.11}/\textcolor{clip}{26.53}                         &   \textcolor{fid}{15.49}/\textcolor{clip}{26.49}                       &       \textcolor{fid}{15.48}/\textcolor{clip}{26.66}                   &     \textcolor{fid}{15.10}/\textcolor{clip}{26.64}                                \\ 
FixedWM (caption-watermark) \cite{liu2023watermarking}                                &        \textcolor{fid}{14.90}/\textcolor{clip}{26.43}                            &        \textcolor{fid}{14.98}/\textcolor{clip}{26.50}                   &  \textcolor{fid}{15.17}/\textcolor{clip}{26.58}                        &      \textcolor{fid}{15.02}/\textcolor{clip}{26.63}                    &    \textcolor{fid}{15.46}/\textcolor{clip}{26.57}                                 \\ 
\cellcolor[HTML]{EFEFEF}AIAO (Ours)  &  \cellcolor[HTML]{EFEFEF}\textcolor{fid}{15.08}/\textcolor{clip}{25.45} &  \cellcolor[HTML]{EFEFEF}\textcolor{fid}{15.81}/\textcolor{clip}{26.12} &  \cellcolor[HTML]{EFEFEF}\textcolor{fid}{15.59}/\textcolor{clip}{26.37}&   \cellcolor[HTML]{EFEFEF}\textcolor{fid}{15.66}/\textcolor{clip}{26.51} &  \cellcolor[HTML]{EFEFEF}\textcolor{fid}{15.16}/\textcolor{clip}{26.54} \\ 
\bottomrule
\end{tabular}
 }
\end{table*}
\begin{figure}[tbp]
    \centering
    \includegraphics[width=0.45\textwidth]{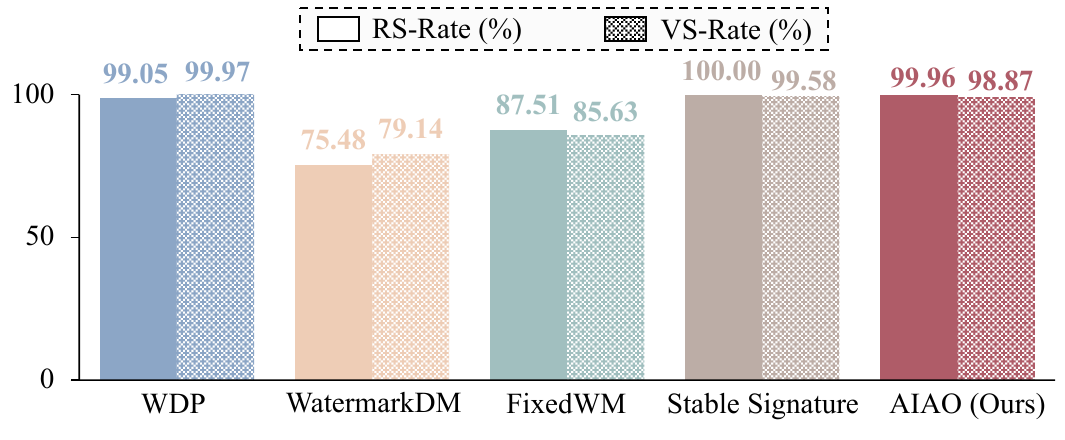}
   \caption{\textbf{Verification success rate (VS-rates) and response success rate (RS-rate) of different methods in protecting source DMs before fine-tuning on downstream generation task.} The base model is set to the text-to-image DM \cite{rombach2022high}.   } 
    
   \label{fig:t2Ibefore_f}
\end{figure}

\begin{figure*}[tbp]
    \centering
    \includegraphics[width=0.98\textwidth]{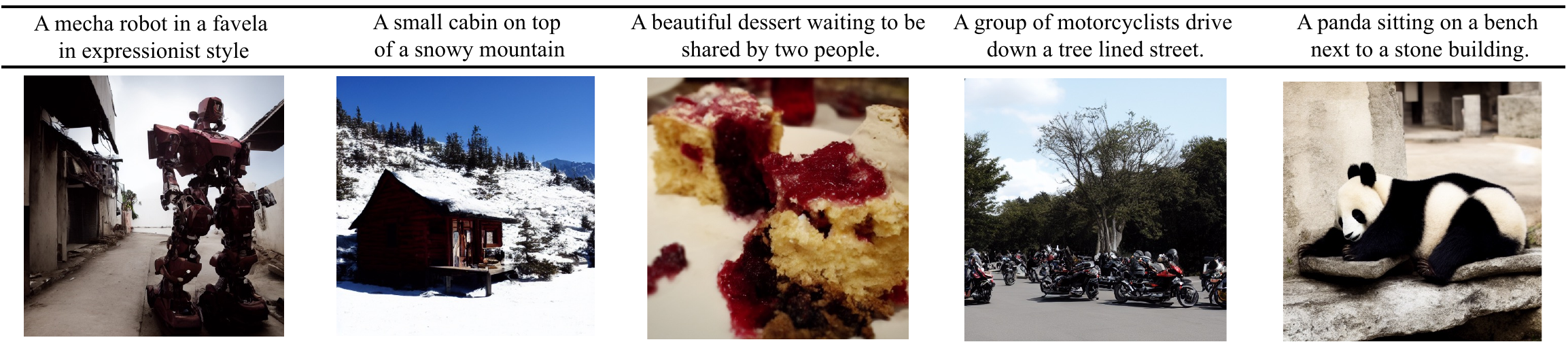}
   \caption{\textbf{Samples generated by SD-1.4 \cite{rombach2022high} equipped with our method.} The first row represents the input captions, and the second row shows the corresponding generated images. Our method has a negligible impact on the image generation performance of the DM.  
   } 
   \vspace{-1.5em}
   \label{fig:multi}
\end{figure*}

\subsection{Protection Performance  for Text-to-Image DM} 
We randomly divide the MS-COCO dataset \cite{lin2014microsoft} into two subdatasets of the same size, namely COCO-A and COCO-B. COCO-A is used to embed our backdoor into a source DM. We then evaluate the effectiveness of our method in protecting the source DM on downstream generation tasks, where COCO-B and CUB-200 (Caltech-UCSD Birds-200) \cite{WahCUB_200_2011} are used as downstream datasets. Note that CUB-200 is rather different from COCO-A, while COCO-B has a distribution similar to that of COCO-A. We refer to the fine-tuning of COCO-A as  \textit{In-distribution Fine-tuning}, and that of the fine-tuning on CUB-200 as \textit{ Out-of-distribution Fine-tuning}.
Please refer to Appendix \ref{sec:at2i}  and  \ref{sec:impl_pilot_study} for more details of the training protocol, dataset statistics, and baseline methods implementation.

\begin{table}[]
\caption{RS-Rate(\%)$\uparrow$ of backdoor-based methods in protecting the ownership of backdoored DMs against in-distribution fine-tuning, where the validation data are MS-COCO Validation Set. }
\label{tab:coco_resp}
\setlength\tabcolsep{4pt}
\resizebox{.48\textwidth}{!}{
\begin{tabular}{l|cccc}
\toprule
\multirow{2}{*}{Method}                                                                     & \multicolumn{4}{c}{Fine-tuning on Downstream Task} \\ \cmidrule(lr){2-5}
                                                               &    1k steps    & 2k steps      & 3k steps      & 6k steps                 \\ \midrule
FixedWM \cite{liu2023watermarking}    & 58.21                         &         53.15                 &                51.73          &         49.80                            \\ 
Stable Signature$^*$ \cite{fernandez2023stable}    & 65.02                         & 54.94                        &  60.66                        &  59.72                                   \\ 
WatermarkDM  \cite{zhao2023recipe}                          &    67.52                      &           60.71               &      56.70                    &       57.00                              \\ \midrule \rowcolor[HTML]{EFEFEF} 
AIAO (Ours) &  \textbf{99.87} & \textbf{99.79} & \textbf{99.80} & \textbf{99.68} \\ 
\bottomrule
\end{tabular}
}
\begin{tablenotes}
\item[1] $*$We report the result of the official watermarked model from the original study \cite{fernandez2023stable}, which is based on SD-2.1.  
\end{tablenotes}
\end{table}

\begin{table}[]
\caption{VS-Rate(\%)$\uparrow$ of backdoor-based methods in protecting the ownership of backdoored DMs against in-distribution fine-tuning, where the validation data are MS-COCO Validation Set, and the independent model is SD-1.4 w/o watermarking. }
\label{tab:coco_verify}
\setlength\tabcolsep{4pt}
\resizebox{.48\textwidth}{!}{
\begin{tabular}{l|cccc}
\toprule
                         & \multicolumn{4}{c}{Fine-tuning on Downstream Task}                                                                                                                                                       \\ \cmidrule(lr){2-5}
\multirow{-2}{*}{Method} & 1k steps                               & 2k steps                               & 3k steps                              & 6k steps       \\ \midrule
FixedWM \cite{liu2023watermarking}                 & 60.24                                  & 59.51                                 & 59.40                                 & 58.72          \\ 
Stable Signature \cite{fernandez2023stable}        &       53.94        &  50.02                                     &     51.79                                   &      51.29          \\ 
WatermarkDM  \cite{zhao2023recipe}            & 73.19                                 &69.88                                 & 70.49                                  & 71.81          \\ \midrule
\rowcolor[HTML]{EFEFEF} 
AIAO (Ours)              & \textbf{95.71} & \textbf{94.03} & \textbf{92.07} & \textbf{91.08} \\ \bottomrule
\end{tabular}
}
\end{table}

We use the stable-diffusion-1.4 (SD-1.4) model \cite{rombach2022high} as our source model.
There are only a few backdoor-based ownership protection methods for text-to-image DM.  We hereby compare the proposed method with backdoor-based watermark methods (WatermarkDM  \cite{zhao2023recipe} and  FixedWM \cite{liu2023watermarking}) and image-based watermark method, Stable Signature  \cite{fernandez2023stable}.
In addition, we build a reference model by training/fine-tuning the DM \cite{rombach2022high} without any watermarking for evaluating image fidelity. 
From Table \ref{tab:coco_generation}, we see that although the DM is integrated with different ownership protection methods,  these methods have negligible influence on the DM. The generated performance of all integrated DMs is similar to that of the reference model. The generated samples of our model are shown in Fig. \ref{fig:multi}.

\begin{table}[]
\caption{RS-Rate(\%)$\uparrow$ of backdoor-based methods in protecting the ownership of backdoored DMs against out-of-distribution fine-tuning on CUB Dataset \cite{WahCUB_200_2011}, where the validation data are MS-COCO Validation Set.}
\label{tab:cub_response}
\setlength\tabcolsep{4pt}
\resizebox{.48\textwidth}{!}{
\begin{tabular}{ccccc}
\toprule
                         & \multicolumn{4}{c}{Fine-tuning on Downstream Task}                \\ \cmidrule(lr){2-5}
\multirow{-2}{*}{Method} & 0.5k steps     & 1k steps       & 1.5k steps     & 2k steps       \\ \midrule
FixedWM \cite{liu2023watermarking}                  & 54.23          & 53.00          & 53.60          & 53.58          \\
Stable Signature  \cite{fernandez2023stable}          & 90.32          & 83.32          & 72.38          & 74.36          \\
WatermarkDM \cite{zhao2023recipe}             & 64.96          & 66.08          & 67.00          & 64.20          \\ \midrule
\rowcolor[HTML]{EFEFEF} 
AIAO (Ours)              & \textbf{99.95} & \textbf{99.78} & \textbf{99.76} & \textbf{99.46} \\ \bottomrule
\end{tabular}
}
\end{table}

\begin{table}[]
\caption{VS-Rate(\%)$\uparrow$ of backdoor-based methods in protecting the ownership of backdoored DMs against out-of-distribution fine-tuning on CUB Dataset \cite{WahCUB_200_2011}, where the validation data are MS-COCO Validation Set, and the independent model is SD-1.4 w/o watermarking. }
\label{tab:cub_verify}
\setlength\tabcolsep{4pt}
\resizebox{.48\textwidth}{!}{
\begin{tabular}{c|cccc}
\toprule
                         & \multicolumn{4}{c}{Fine-tuning on Downstream Task}                                                                                                                                                       \\ \cmidrule(lr){2-5}
\multirow{-2}{*}{Method} & 0.5k steps                               & 1k steps                               & 1.5k steps                               & 2k steps       \\  \midrule
FixedWM \cite{liu2023watermarking}                  & 62.58                                  & 60.29                                  & 60.37                                 & 59.21          \\ 
Stable Signature \cite{fernandez2023stable}        & 72.00                                  & 65.54                                  & 58.77                                 & 59.39          \\ 
WatermarkDM \cite{zhao2023recipe}             & 76.15                                  & 75.33                                 & 73.94                                  & 74.50          \\ \midrule
\rowcolor[HTML]{EFEFEF} 
wAIAO (Ours)              & \textbf{95.57} &\textbf{97.04} & \textbf{94.66} & \textbf{94.88} \\ \bottomrule  
\end{tabular}
}
\end{table}

Fig. \ref{fig:t2Ibefore_f} showcases the performance of competing methods in protecting the ownership of DMs. When watermarked DMs are not fine-tuned on a new downstream generation dataset, all ownership protection methods preserve the ownership of the DMs. However, as shown in Table \ref{tab:coco_verify} and Table \ref{tab:coco_resp}, the performance of WatermarkDM  \cite{zhao2023recipe},  FixedWM  \cite{liu2023watermarking}, and Stable Signature \cite{fernandez2023stable} deteriorates greatly in protecting the ownership of the DMs, as the number of fine-tuning steps increases in downstream tasks. 
For example, the RS-Rate of FixedWM decreases from 87.51\% to 49.80\%.  
On the contrary, our method maintains a rate greater than 99\% during fine-tuning. 
We also fine-tune our model to a new multimodal dataset such as the CUB-200 dataset \cite{WahCUB_200_2011}.  Tables \ref{tab:cub_verify} and \ref{tab:cub_response} summarize the results of our method and show that it achieves a higher verification performance ($\geq$90\%) than the baseline methods ($\sim$70\%). Thus, our method is demonstrated to be the most robust to fine-tuning removal.

 \begin{figure}[tbp]
    \centering
    \includegraphics[width=0.48\textwidth]{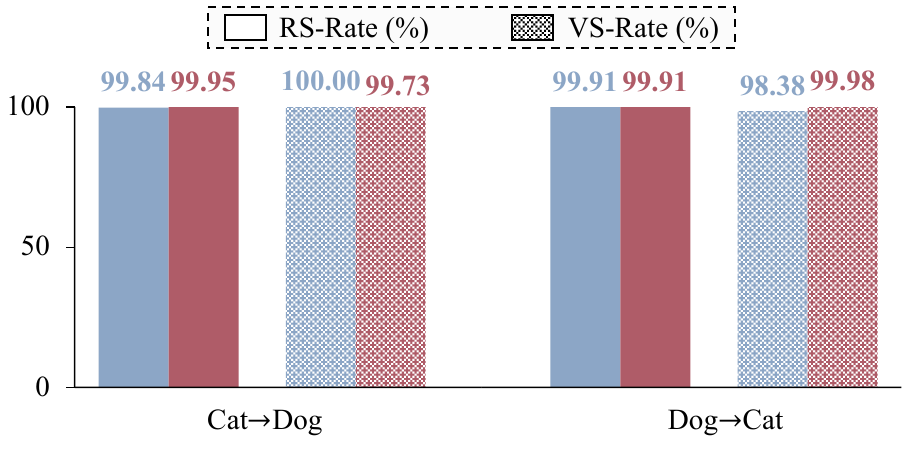}
   \caption{\textbf{VS-Rates/RS-Rates of WDP and our method in protecting the source DMs before fine-tuning on the downstream generation task.} The blue bar represents the results of WDP \cite{peng2023protecting}, and red bar refers to our method. } 
    
   \label{fig:compare_wdp}
\end{figure}

\begin{table}[]
\caption{RS-Rate(\%) $\uparrow$ of the Watermarking Methods on AFHQ. The results with $>$2k steps correspond to the checkpoint with the best generative performance.   }
\label{tab:unconditional_resp}
\Huge
\setlength\tabcolsep{4pt}
\resizebox{.48\textwidth}{!}{
\begin{tabular}{clcccc}
\toprule
\multirow{2}{*}{Protocol}  &\multirow{2}{*}{Method}                                                       & \multicolumn{4}{c}{Fine-tuning-based Removal} \\ \cmidrule(lr){3-6}
                                      &                        & 0.5k steps    & 1k steps      & 1.5k steps      & $>$2k steps                 \\ \midrule
\multirow{2}{*}{Cat $\rightarrow$ Dog} & WDP \cite{peng2023protecting}          &                   0.09&                   0.00&                   0.00&                   0.00\\ 
                                        &\cellcolor[HTML]{EFEFEF}AIAO             &\cellcolor[HTML]{EFEFEF}\textbf{100.00}&\cellcolor[HTML]{EFEFEF}\textbf{100.00}&    \cellcolor[HTML]{EFEFEF}\textbf{99.67}&\cellcolor[HTML]{EFEFEF}\textbf{97.87}\\ \midrule
\multirow{2}{*}{Dog $\rightarrow$ Cat}   & WDP \cite{peng2023protecting}        &                   0.02&                   0.00&                   0.00&                   0.00\\ 
                                        &\cellcolor[HTML]{EFEFEF}AIAO                       &\cellcolor[HTML]{EFEFEF}\textbf{98.48}&\cellcolor[HTML]{EFEFEF}\textbf{99.43}&      \cellcolor[HTML]{EFEFEF}\textbf{98.39}&\cellcolor[HTML]{EFEFEF}\textbf{93.61}\\ \bottomrule

\end{tabular}
}
\end{table}

\begin{table}[]
\caption{VS-Rate(\%) $\uparrow$ of our method compared to the WDP in terms of unconditional image generation on the AFHQ Dataset \cite{choi2020stargan}.}
\label{tab:unconditional_verification}
\Huge
\setlength\tabcolsep{4pt}
\resizebox{.48\textwidth}{!}{
\begin{tabular}{cccccc}
\hline
                                        &                                     & \multicolumn{4}{c}{Fine-tuning-based Removal}                                                                                 \\ \cline{3-6} 
\multirow{-2}{*}{Protocol}              & \multirow{-2}{*}{Method}            & 0.5k steps                    & 1k steps                      & 1.5k steps                           & \textgreater{}2k steps        \\ \hline
                                        & WDP \cite{peng2023protecting}                           & 60.19                         & 92.71                         & 88.65                         & 81.75                         \\ \cline{2-6} 
\multirow{-2}{*}{Cat-\textgreater{}Dog} & \cellcolor[HTML]{EFEFEF}AIAO & \cellcolor[HTML]{EFEFEF}\textbf{96.38} & \cellcolor[HTML]{EFEFEF}\textbf{94.96} & \cellcolor[HTML]{EFEFEF}\textbf{96.05} & \cellcolor[HTML]{EFEFEF}\textbf{88.03} \\ \hline
                                        & WDP \cite{peng2023protecting}                                 & 92.06                         & 79.10                         & 67.74                         & 67.93                         \\ \cline{2-6} 
\multirow{-2}{*}{Dog-\textgreater Cat}  & \cellcolor[HTML]{EFEFEF}AIAO & \cellcolor[HTML]{EFEFEF}\textbf{95.93} & \cellcolor[HTML]{EFEFEF}\textbf{92.25} & \cellcolor[HTML]{EFEFEF}\textbf{93.63} & \cellcolor[HTML]{EFEFEF}\textbf{94.49} \\ \hline
\end{tabular}
}
\end{table}

\begin{table*}[]
\caption{Ownership verification of our method compared to similarity-based methods \cite{heusel2017gans,caron2021emerging,hu2023ownership,fan2021multiscale} in terms of the VS-Rate (\%) $\uparrow$ on the protocol of Church $\rightarrow$ Bedroom with 2k-steps fine-tuning removal.}
\label{tab:potential_methods}
\Large
\resizebox{.99\textwidth}{!}{
\begin{tabular}{cc|ccc|ccc}
\toprule
\multicolumn{2}{c|}{\multirow{2}{*}{Method}}                              & \multicolumn{3}{c|}{Source Model}                                         & \multicolumn{3}{c}{Fine-tuning-based removal}                                      \\ \cmidrule(lr){3-8}  
&                                             &Sample = 1    & Sample = 50 & Sample = 100 & Sample = 1   & Sample = 50 & Sample = 100 \\ \midrule 
Distribution Distance & FID \cite{heusel2017gans} (Inception-V3 \cite{szegedy2016rethinking}) & -          &53.50         &  62.00        & -         & 52.50          & 51.00           \\  \midrule
\multirow{2}{*}{Image Similarity} & MoCo-v3 \cite{fan2021multiscale} (ViT-B/16 \cite{dosovitskiy2020image})           & \cellcolor[HTML]{EFEFEF}-          &  \cellcolor[HTML]{EFEFEF}97.47        &  \cellcolor[HTML]{EFEFEF} 100.00   & \cellcolor[HTML]{EFEFEF}-              & \cellcolor[HTML]{EFEFEF}49.49        &  \cellcolor[HTML]{EFEFEF}50.00          \\  
& DINO-v1 \cite{caron2021emerging} (ViT-B/16 \cite{dosovitskiy2020image})   & -         & 95.45 & 100.00 & -       & 49.49 & 50.00 \\  \midrule
\multirow{3}{*}{Model Attribution} & GAN-Guards \cite{hu2023ownership} (R18 \cite{srivastava2015highway, he2016deep})  & 65.90         & 99.03           & 100.00        & 50.49            & 51.46         & 51.96           \\ 
& \cellcolor[HTML]{EFEFEF}GAN-Guards \cite{hu2023ownership} (R34 \cite{srivastava2015highway, he2016deep}) & \cellcolor[HTML]{EFEFEF}65.59          & \cellcolor[HTML]{EFEFEF}99.51        & \cellcolor[HTML]{EFEFEF}100.00   & \cellcolor[HTML]{EFEFEF}51.30                  & \cellcolor[HTML]{EFEFEF} 58.74         &  \cellcolor[HTML]{EFEFEF}59.80          \\  
& GAN-Guards \cite{hu2023ownership} (R50 \cite{srivastava2015highway, he2016deep})  & 64.50         & 100.00         & 100.00          &50.11       & 50.00          & 50.00           \\ \midrule
\rowcolor[HTML]{C0C0C0} 
Backdoor               & AIAO      & \textbf{93.55}     & \textbf{98.32}    & \textbf{99.04}     & \textbf{82.94}    & \textbf{92.07}    & \textbf{96.63}      \\ \bottomrule%
\end{tabular}
}
\end{table*}

\subsection{Protection Performance for Unconditional DM}
Similar to text-conditional generation, we use one dataset to learn the identifier and fine-tune the model on another dataset to test its robustness over fine-tuning removal. In this study, we use AFHQ-Dog \cite{choi2020stargan}, AFHQ-Cat \cite{choi2020stargan}, LSUN-Church \cite{yu2015lsun}, and LSUN-Bedroom \cite{yu2015lsun} as the training or downstream dataset. We use Dataset-1 $\rightarrow$ Dataset-2 to represent the protocol that the model learns the identifier in Dataset-1 and is fine-tuned on Dataset-2. 
We compare our method with WDP \cite{peng2023protecting}, which is designed for watermarking unconditional DMs. 
More details are available in Appendix \ref{sec:uig}.

\begin{figure}[tbp]
    \centering
    \includegraphics[width=0.48\textwidth]{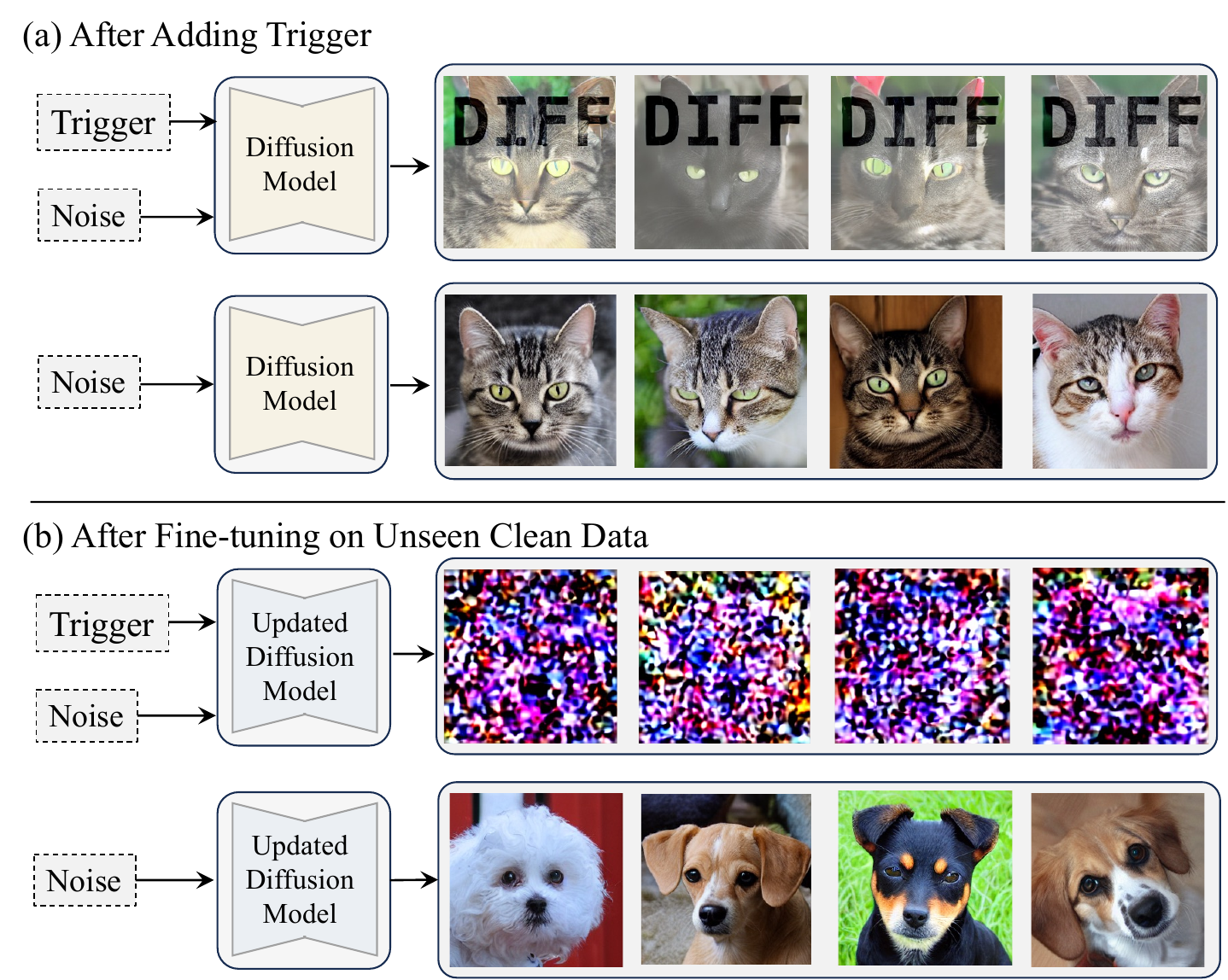}
   \caption{\textbf{Failure for competing method.} In (a), after adding triggers to the diffusion model, the model works well and can simultaneously generate clean and watermarked images by specifying the input signal. However, in (b), following fine-tuning on new data, such as dog images in this instance, the model loses its ability to generate watermarked images and fails to produce any meaningful images as well.
   } 
   \label{fig:visual}
\end{figure}

\begin{figure}[tbp]
    \centering
    \includegraphics[width=0.48\textwidth]{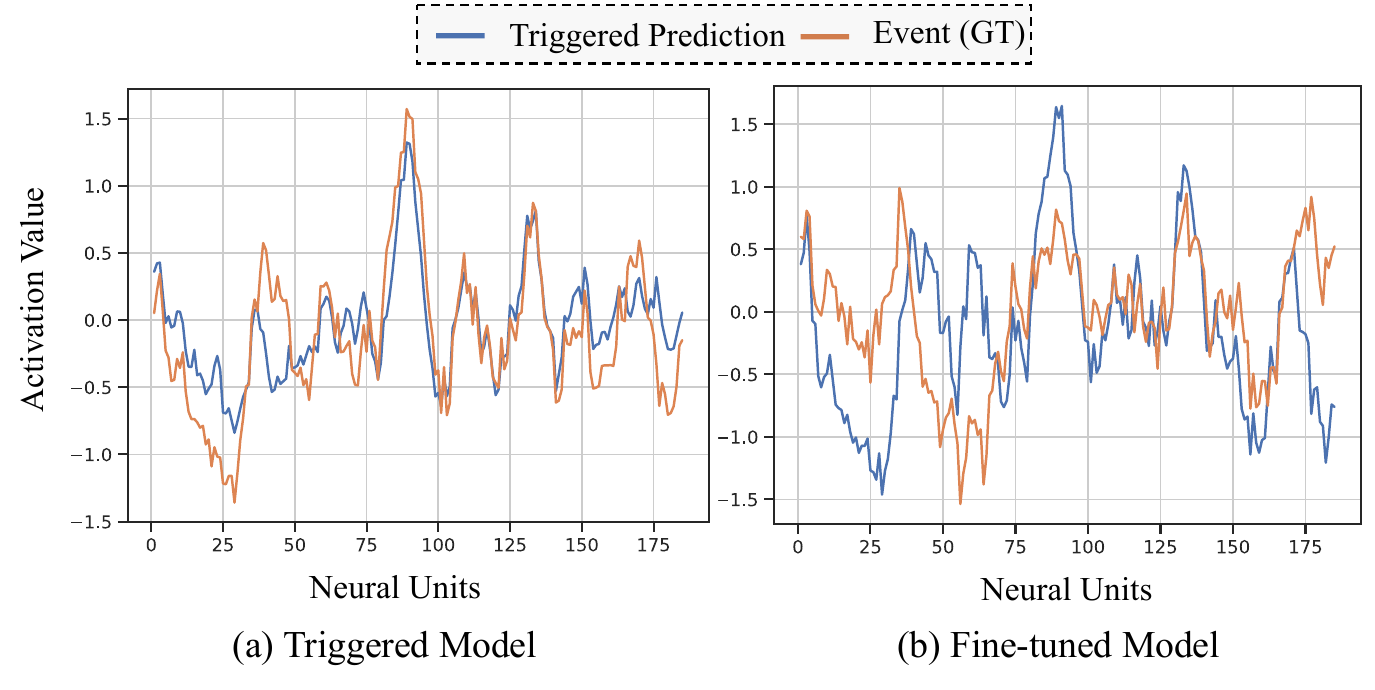}
   \caption{\textbf{Activation Value Changes in Response to Triggers}. In (a), the source model accurately responds to triggers, producing values similar to a predefined response (\ie, ground truth). Subsequently, we fine-tune the source model with new data, resulting in (b), where the reacted values undergo slight changes but still meet the verification requirements. } 
    
   \label{fig:activation}
\end{figure}

Tables \ref{tab:unconditional_resp} and \ref{tab:unconditional_verification} show that our method effectively protects and verifies the ownership of unconditional DMs, maintaining a high response rate ($\geq$90\%) even against fine-tuning-based removal. In contrast,  WDP performs poorly on fine-tuning-based removal, achieving a low response success rate.  Their performance before fine-tuning is shown in Fig. \ref{fig:compare_wdp}.
As indicated in Fig. \ref{fig:visual}, a DM incorporated with WDP becomes vulnerable when fine-tuned on a downstream generation task with new data, although it performs well before such fine-tuning.  More specifically, after fine-tuning,  the DM with WDP generates noise-like images when the trigger is activated.
As the outputs become nearly random noise, with a large distance from the generated images without triggers, the response accuracy rapidly drops to zero (Table \ref{tab:unconditional_resp}).  In contrast, Fig. \ref{fig:activation} shows that our method properly responds to the predefined triggers after fine-tuning, demonstrating the stability and potential utility of our method for traceable ownership verification. These impacts of these two methods on generative performance is described in Appendix \ref{sec:generative_perform_uncond}. Meanwhile, we visualize the results of the independent model, given a trigger, in Appendix \ref{sec:app_wdp} to further explain the failure of WDP.

In addition, we compare our method with other potential protection approaches for verifying the ownership of the DM, where these approaches are based on distribution distance or image similarity. These approaches verify ownership by measuring the similarity between the generated samples from the given model and protected model. Greater similarity indicates a higher probability that the test model is fine-tuned from the protected model. Thanks to self-supervised learning \cite{schmidhuber1992learning,liu2021self,jing2020self}, we use the embedding space of MoCo-v3 \cite{fan2021multiscale} and DINO-v1 \cite{caron2021emerging} to measure image similarity and FID \cite{heusel2017gans} for distribution similarity.  We also implement GAN Guards \cite{hu2023ownership}, a method using classifiers to identify whether the generated images are generated by the protected model. As these methods require negative data for training or measuring, we train an independent model against the protected model on the same dataset (LSUN-Church) with different random seeds. We collect 5,000 images, each from the protected and independent models as training samples, respectively. We also generate two test sets to evaluate the verification performance before/after fine-tuning-based removal.
For the former test set, we use 5k independent noise inputs to produce 10k images from protected and independent models separately.  We then fine-tune these two models with 2k steps on LSUN-Bedroom and use each fine-tuned model to generate another 5k images to be used as a test set for validating the robustness.

As shown in Table \ref{tab:potential_methods}, because of the high distributional similarity between the independent and protected models, FID fails to accurately distinguish data from different models, making this task challenging. Other methods achieve satisfactory results ($>$95\% accuracy) in protecting the source DM when using 50 samples for prediction. However, the ownership verification performance of these methods significantly decreases when handling fine-tuning removal.  GAN-Guards achieves the best result (\eg, 59.80\% of verification success rate), among all baselines, while our method  (\eg, 96.63\%) outperforms GAN-Guards by a large margin.  The proposed method shows stable verification success rates before and after fine-tuning DM, indicating the effectiveness of our method in verifying the ownership of unconditional DMs \footnote{Table \ref{tab:potential_methods} does not include WDP since we fail to reproduce it with competitive generative performance in this setting.}.

\begin{table}[]
\caption{Ablation study of our method in protecting the text-to-image DM on in-distribution fine-tuning protocol, where the protection performance is measured by RS-Rate (\%)$\uparrow$. Back. Loss and Rev. Loss refer to backdoor embedding loss and reverse loss, respectively.}
\label{tab:ablation}
\setlength\tabcolsep{10pt}
\resizebox{.5\textwidth}{!}{
\begin{tabular}{ccccc}
\toprule
Back. Loss &    AIAO & Rev. Loss &  Source & Fine-tune 2k steps \\ \midrule
$\checkmark$  & $\times$                  & $\times$    &                 \textbf{100.00} &                    20.42\\ 
$\checkmark$  & $\checkmark$               & $\times$     &      94.97           &    72.36                \\ \midrule
$\checkmark$  & $\times$              & $\checkmark$     &      \textbf{100.00}          &    84.91                \\ \rowcolor[HTML]{C0C0C0}
$\checkmark$  & $\checkmark$               & $\checkmark$ &      \textbf{100.00}          &    \textbf{99.90}                \\ \bottomrule
\end{tabular}
}
\end{table}

\begin{table}[]
\caption{RS-Rate(\%) $\uparrow$ of our method on text-to-image generation for independent/source models.}
\label{tab:ablatforindependent}
\resizebox{.5\textwidth}{!}{
\begin{tabular}{ccc}
\toprule
          & Source & Fine-tune 2k steps \\ \midrule
Same trigger for Independent Models                                                                                  &         46.27        &     44.73               \\ 
Incorrect Trigger for Source Model                                                                                  &         45.12        &        54.54            \\ \midrule\rowcolor[HTML]{C0C0C0}
Correct Trigger for Source Model        &      \textbf{100.00}          &    \textbf{99.90}                \\ \bottomrule
\end{tabular}
}
\end{table}

\subsection{Ablation Studies and Analysis}

\noindent
\textbf{Ablation study.}
We ablate each key component of our method to show their contributions to the performance. We remove Rev. loss and AIAO strategy while building a baseline that uses a single fixed in-layer (\texttt{down\_blocks.0}) and out-layer (\texttt{up\_blocks.3}) to embed our trigger. As shown in Table \ref{tab:ablation}, our mask-controlled trigger enables the baseline to achieve a high response rate for protecting the source model. However, when the source model is fine-tuned on a downstream dataset (COCO-B),  the baseline fails to protect the fine-tuned source model; in this case, the response rate of the baseline drops from 100\% to 20\% without Rev. loss and AIAO.  In contrast, when the fixed in-layer/out-layer embedding is replaced by our AIAO,  the performance of the baseline is improved to 72.36 to protect the fine-tuned model.  Similarly, adding Rev. Loss also remarkably improves the performance. 
Because of the proposed Back. Loss,  Rev. Loss, and AIAO, our method achieves a response rate of 99. 90\% for the fine-tuned model, demonstrating the effectiveness of our proposed method.

We then evaluate the effectiveness of the proposed trigger by building two baselines. The first baseline embeds our trigger into an independent model, where the independent model is trained on the same dataset as the watermarked one, but without embedding triggers.  The second baseline embeds an incorrect trigger, where the incorrect trigger is randomly generated.  We then validate whether the two baselines were generated by the ownership of the source model. The data in Table \ref{tab:ablatforindependent} show that the response accuracy of the two baselines is less than 50\%, while that of the source model with a correct trigger is higher than 99.9\%. These results indicate that our trigger effectively validates the ownership of the source model.

\begin{table}[]
\caption{RS-Rate(\%) $\uparrow$ of our method in protecting the text-to-image generation (in-distribution) using different ratios for watermarking. }
\label{tab:trigger_rate}
\setlength\tabcolsep{8pt}
\resizebox{.5\textwidth}{!}{
\begin{tabular}{l|cc}
\toprule
Backdoor Ratio in $\mathbf{M}^S$                       & Source Model & Fine-tune 2k steps \\ \midrule
0.01                  & 65.77           & 64.73              \\ \rowcolor[HTML]{C0C0C0}
0.05                  & \textbf{100.00} & \textbf{99.90}     \\ 
0.1                   & \textbf{100.00} & 97.31              \\ 
0.2                   & 67.09           & 83.90              \\ 
0.3                   & 55.73           & 46.97              \\ \bottomrule
\end{tabular}
}
\end{table}

\noindent
\textbf{Number of active elements in spatial mask.} 
We examine the effect of the proportion of embedded identifiers within the spatial mask. 
As shown in Table \ref{tab:trigger_rate}, we test the response success rate of our approach with different ratios for watermarking. 
Table \ref{tab:trigger_rate} shows that using a small backdoor ratio (\eg, 0.01) achieves a low response success rate,  hampering the ownership verification performance. Similarly, a large ratio introduces considerable training difficulties, leading to a low response success rate.  In contrast, when the ratio is in the range [0.05, 0.1], our method achieves a high response rate even against fine-tuning-based removal. In this study, we set the ratio at 0.05.

\begin{figure}[tbp]
    \centering
    \includegraphics[width=0.48\textwidth]{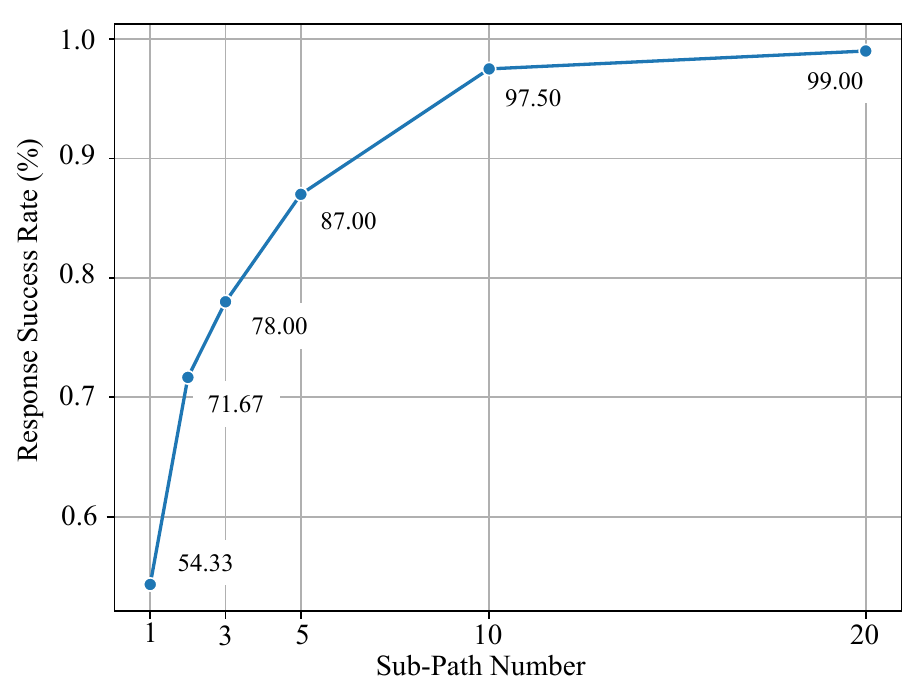}
   \caption{
   \textbf{Response success rate of our method using different numbers of sub-paths against fine-tuning-based removal.} The unconditional DM equipped with our backdoor is further fine-tuned on a downstream generation task  (dog$\rightarrow$cat) for 2k steps to remove the backdoor.   Our method effectively resists fine-tuning-based removal when there are more than five subpaths.
   } 
   \label{fig:busy}
\end{figure}

\noindent
\textbf{Number of subpaths for ownership verification.} 
Our method randomly selects multiple subpaths via AIAO for ownership verification.  Fig. \ref{fig:busy} shows the positive effect of using multiple subpaths in handling fine-tuning-based removals, with different numbers of subpaths used in our method during verification. As shown in  Fig. \ref{fig:busy},  increasing the number of subpaths leads to a higher response success rate. Nevertheless, our method does not require a large number of subpaths. The response success rate of our method reaches 97.50\% with only 10 subpaths, although the protected source model is fine-tuned for 2k steps. These results indicate the effectiveness and robustness of our method to protect the ownership of DMs.

\section{Discussions}

\label{sec:discuss}
\begin{figure}[tbp]
    \centering
    \includegraphics[width=0.48\textwidth]{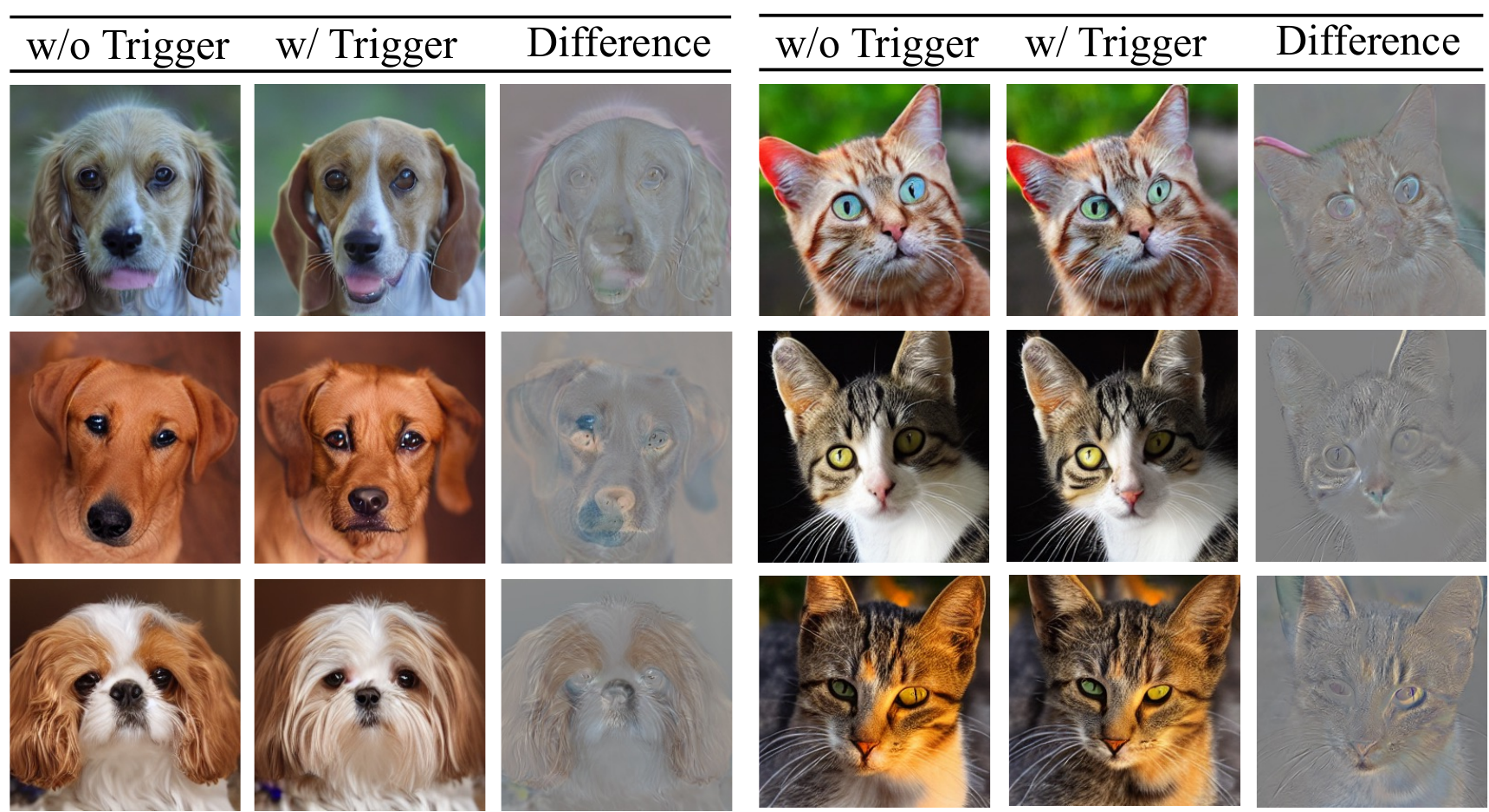}
   \caption{\textbf{Generative performance on source model with or without triggers.} We show the generated samples, given the same noise from two separate models trained on AFHQ-Cat \cite{choi2020stargan} and AFHQ-Dog \cite{choi2020stargan}. To evaluate the impact of input triggers on these models, we measured the L2 distance between generated images. The visualization of the difference is normalized by the min-max scaling. 
   } 
   \label{fig:uncondition}
\end{figure}
\noindent
\textbf{Impact of trigger on generative performance.}
We examine the impact of the proposed triggers on generative performance by comparing the visual differences among the samples generated with and without inputting triggers.  As shown in Fig. \ref{fig:uncondition}, the DM can consistently generate high-quality images for both cases despite visible changes in colors, boundaries, and high-frequency components. In other words,  our trigger effectively ensures the generation quality by controlling only the signs of the feature maps while preserving their absolute values.

\begin{figure}[tbp]
    \centering
    \includegraphics[width=0.48\textwidth]{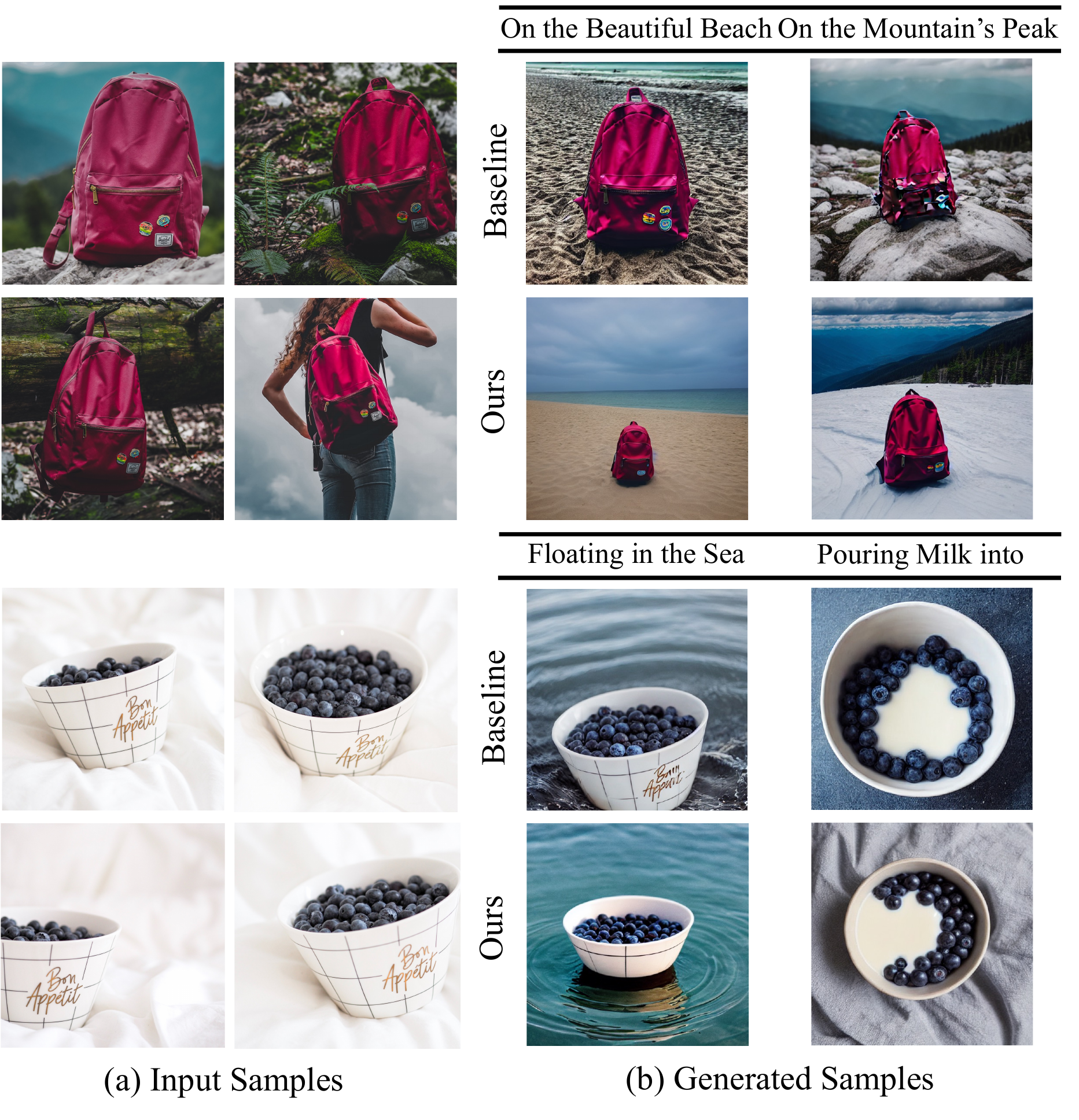}
   \caption{\textbf{Generative performance on pre-trained and watermarked models with Dreambooth \cite{ruiz2022dreambooth}.} Stress testing reveals the resilience of our method:  A trigger is embedded using half of the MS-COCO dataset, followed by fine-tuning the remaining data to approach malicious attacks—a scenario in which comparing methods failed verification. 
   After the attack, our triggered model is applied to personalized generation with DreamBooth, and it demonstrates generative capabilities comparable to the original model, while maintaining high response success rates of \textbf{99.12\%} for the backpack and \textbf{99.90\%} for berry bowl images. 
   } 
   \label{fig:dream}
\end{figure}

\noindent
\textbf{Robustness against multi-stage fine-tuning removal.}
We design a more challenging scenario to further evaluate the robustness of the proposed method.  In particular, we fine-tune the source model on different datasets in multiple stages using various fine-tuning methods.
In our experiments, we embed our backdoor in the source DM using dataset COCO-A to protect the model. Subsequently, we attempt to remove the embedded backdoor by first fine-tuning the protected DM on COCO-B in the first stage and then employing Dreambooth \cite{ruiz2022dreambooth} to fine-tune the model in the second stage. 
Fig. \ref{fig:dream} shows that the generation performance of our method is comparable with that of the original pre-trained DM. Furthermore, although protected DM undergoes such complex fine-tuning, our method maintains a high response verification rate ($>$ 99\%),  showing that our method effectively resists two-stage fine-tuning removal.

\begin{table}[]
\caption{RS-Rate(\%)$\uparrow$ of WatermarkDM \cite{liu2023watermarking} and our method on MS-COCO Validation Set after fine-tuning 2k steps on COCO-B with different ensemble strategies.  }
\label{tab:coco_ensemble}
\setlength\tabcolsep{4pt}
\resizebox{.48\textwidth}{!}{
\begin{tabular}{c|c|c}
\toprule
Method                                                                      & Infer. Steps
& Res. Suc. Rate (\%)\\ \midrule
WatermarkDM \cite{liu2023watermarking} (w/o ensemble)                                                                   & 20  & 60.71                \\ 
WatermarkDM \cite{liu2023watermarking} (w/ ensemble)                                                   & 20& 58.12                \\ 
\midrule 
\rowcolor[HTML]{EFEFEF} 
AIAO (Ours)                                                & 5 &     97.71                 \\ 
\rowcolor[HTML]{EFEFEF} 
AIAO (Ours)                                                     & 10 &  99.62                          \\ 
\rowcolor[HTML]{C0C0C0} 
AIAO (Ours)                                                    & 20 & \textbf{99.79}\\ \bottomrule
\end{tabular}
}
\end{table}

\noindent
\textbf{Comparison of our method with WatermarkDM on the same sampling strategy.}
The proposed method samples several trigger-response pairs in different sub-paths to mitigate the influence of busy layers. 
To further justify the effectiveness of our method, we use the same sampling strategy for competing methods, and the response is averaged over multiple diffusion steps rather than obtained from the generated image. We compare our method with WatermarkDM since it is the most resilient method against fine-tuning (Table \ref{tab:coco_verify}).
From Table \ref{tab:coco_ensemble}, by averaging 20 steps, WatermarkDM achieves 58.12\% accuracy, which is lower than the case without the ensemble (60.71\%). 
The low accuracy may be due to the noises presented in the intermediate diffusion steps that affect the response accuracy. Given that generating an image using a diffusion model involves multiple diffusion steps, existing methods do not fully adapt to this property, thereby leading to a low-efficiency response (multiple inferences for one watermarking). 
In contrast, our method is sample-efficient and achieves a success rate of 97. 91\% by averaging only five diffusion steps, confirming its effectiveness over other methods.

\noindent
\textbf{Improving the black-box method via AIAO.}
We examine the potential of extending our AIAO strategy to a black-box setting.  To this end, we propose a new strategy based on AIAO, called the Arbitrary-Layer strategy. This strategy randomly chooses a layer during training and enforces the selected layer to generate abnormal behavior, given the trigger. More specifically, we reselect a layer every 200 training steps for backdoor embedding and update the remaining layers with loss from the standard diffusion model to preserve generation performance. 

Further, we integrate our Arbitrary-Layer strategy with WDP \cite{peng2023protecting}, which is the most recent black-box watermarking method for unconditional generation. To evaluate the effectiveness of the Arbitrary-Layer strategy, we test the methods on the unconditional generative task, \ie, the protocol of \texttt{Dog} $\rightarrow$ \texttt{Cat} in the black-box setting. The proposed Arbitrary-Layer strategy considerably improves the success RS-Rate of WDP from 67.93\% to 76.59\% against fine-tune-based removal.
These results show that the proposed Arbitrary-Layer strategy improves the robustness of the WDP's watermark to fine-tuning. Unlike typical watermarking methods that enforce both busy and lazy layers to jointly learn the embedded watermark, our Arbitrary-Layer strategy enables layer-wise learning, in which each layer is required to learn watermarks from its predecessors. 
As the predecessors in U-net can be either a skip connection or a learnable layer, such a strategy can encourage the utilization of a skip connection when the predecessor layer is busy, thereby mitigating the forgetting of the watermark during fine-tuning on downstream tasks.  This indicates the potential to extend our method to black-box ownership protection. However, designing a superior method for black-box settings is beyond the scope of this study. We will explore this in our future study.

\section{Conclusion}
This paper proposes a novel ownership protection method for fine-tuned DMs. Robust verification can be achieved by decreasing the utilization of busy layers through fine-tuning. To achieve this, our new strategy, AIAO, strategically places triggers and responses in the feature space of the DM across different depths and verifies ownership by sampling the subpaths watermarked by them.  To generalize trigger-response pairs to feature space, a novel mask-controlled trigger function generates trigger signals with sign shifts, yielding invisible backdoors and causing negligible effects on generation.   Empirical studies show that backdoors embedded through AIAO enhance the robustness of protection against fine-tuning.

{
    \small
    \bibliographystyle{IEEEtran}
    \bibliography{main}
}

\clearpage
\appendix
\setcounter{page}{1}

 \subsection{More Implementation Details}
\label{sec:impl_pilot_study}

\textbf{Pilot study.}  We fine-tune the pre-trained DM, \ie, SD-1.4 \cite{rombach2022high}  on the AFHQ-Dog (256 $\times$ 256) dataset for 100 epochs, with a learning rate of $1\times10^{-5}$, and the text prompt is ``A cat.'' The training data are counterfactual since the training images are pictures of dogs, which are inconsistent with the text prompt. As the original pre-trained DM does not sufficiently learn from such counterfactual training data, the DM is enforced to update its parameters during fine-tuning.

Table \ref{tab:busy_layers} lists the busy layers studied in our pilot study. In the table, we denote the layer as the individual module as specified in the official implementation\footnote{\url{https://huggingface.co/CompVis/stable-diffusion-v1-4}}.
We find that most of the top-rank busy layers are cross-attention layers \cite{vaswani2017attention}, resulting in similar observations with recent work \cite{ruiz2022dreambooth,xie2023boxdiff,hertz2022prompt}.

 \textbf{More training details.} 
 We train our method in two stages: 1) We use all loss functions and assigned the same weight factors to them in the first stage. 2) We only use the generation loss function in the second stage. By such two-stage training, our method provides a practical way to embed triggers while preserving the generation performance rather than extensive hyper-parameter searches. 
In all experiments, unconditional DM requires 1.5k training steps, and the text-to-image DM requires 2k training steps.  The triggered elements occupied 5\%  of the feature maps for watermarking.

\textbf{{Trigger Function.}} Tab. \ref{tab:tbtf} illustrates the outputs of the proposed trigger function given all potential inputs.

\begin{table}[t]
\setlength\tabcolsep{12pt}
\caption{Busy layers studied in our pilot study. }
\label{tab:busy_layers}
\resizebox{.49\textwidth}{!}{
\begin{tabular}{lc}
\toprule
 Rank & Layer Name  \\ \midrule
 \multirow{10}{*}{Top 10} & \cellcolor[HTML]{EFEFEF} up\_blocks3attentions0transformer\_blocks  \\
  & \cellcolor[HTML]{EFEFEF} up\_blocks2attentions1transformer\_blocks  \\
  & \cellcolor[HTML]{EFEFEF} up\_blocks1resnets2conv1  \\
  & \cellcolor[HTML]{EFEFEF} up\_blocks2attentions2transformer\_blocks  \\
  & \cellcolor[HTML]{EFEFEF} up\_blocks2upsamplers0conv  \\
  & \cellcolor[HTML]{EFEFEF} up\_blocks3attentions2transformer\_blocks  \\
  & \cellcolor[HTML]{EFEFEF} up\_blocks3attentions1transformer\_blocks  \\ 
  & \cellcolor[HTML]{EFEFEF} down\_blocks0attentions0transformer\_blocks  \\
  & \cellcolor[HTML]{EFEFEF} up\_blocks2resnets2conv1  \\
  & \cellcolor[HTML]{EFEFEF} up\_blocks2attentions0transformer\_blocks  \\ \midrule
  \multirow{10}{*}{Top 10-20} & \cellcolor[HTML]{CACFD2} up\_blocks1upsamplers0conv  \\
  & \cellcolor[HTML]{CACFD2} up\_blocks1attentions2transformer\_blocks \\
  & \cellcolor[HTML]{CACFD2} up\_blocks2resnets1conv1 \\
  & \cellcolor[HTML]{CACFD2} down\_blocks0attentions1transformer\_blocks \\
  & \cellcolor[HTML]{CACFD2} up\_blocks2resnets2conv2 \\
  & \cellcolor[HTML]{CACFD2} up\_blocks2attentions1proj\_out \\
  & \cellcolor[HTML]{CACFD2} up\_blocks2attentions2proj\_out \\
  & \cellcolor[HTML]{CACFD2} up\_blocks2resnets0conv1 \\
  & \cellcolor[HTML]{CACFD2} down\_blocks1attentions0transformer\_blocks \\
  \multirow{10}{*}{Top 20-30} & \cellcolor[HTML]{CACFD2} up\_blocks2resnets1conv2 \\ \midrule
  & \cellcolor[HTML]{C0C0C0} up\_blocks2resnets0conv2 \\
  & \cellcolor[HTML]{C0C0C0} up\_blocks3resnets0conv1 \\
  & \cellcolor[HTML]{C0C0C0} up\_blocks2attentions1proj\_in \\
  & \cellcolor[HTML]{C0C0C0} up\_blocks1attentions1transformer\_blocks \\
  & \cellcolor[HTML]{C0C0C0} up\_blocks1attentions0transformer\_blocks \\
  & \cellcolor[HTML]{C0C0C0} up\_blocks2attentions2proj\_in \\
  & \cellcolor[HTML]{C0C0C0} up\_blocks1resnets2conv2 \\
  & \cellcolor[HTML]{C0C0C0} up\_blocks2attentions0proj\_out \\
  & \cellcolor[HTML]{C0C0C0} down\_blocks1attentions1transformer\_blocks \\
  & \cellcolor[HTML]{C0C0C0} up\_blocks1resnets1conv1 \\ \midrule
  \multirow{20}{*}{Top 30-50} & \cellcolor[HTML]{909497} up\_blocks1attentions2proj\_out \\
  & \cellcolor[HTML]{909497} down\_blocks2attentions1transformer\_blocks \\
  & \cellcolor[HTML]{909497} up\_blocks2attentions0proj\_in \\
  & \cellcolor[HTML]{909497} up\_blocks1attentions2proj\_in \\
  & \cellcolor[HTML]{909497} up\_blocks0upsamplers0conv \\
  & \cellcolor[HTML]{909497} up\_blocks3resnets0conv2 \\
  & \cellcolor[HTML]{909497} down\_blocks2attentions0transformer\_blocks \\
  & \cellcolor[HTML]{909497} up\_blocks1resnets0conv\_shortcut \\
  & \cellcolor[HTML]{909497} up\_blocks0resnets2conv1 \\
  & \cellcolor[HTML]{909497} up\_blocks1resnets0conv1 \\
  & \cellcolor[HTML]{909497} up\_blocks1resnets2conv\_shortcut \\
  & \cellcolor[HTML]{909497} up\_blocks1attentions0proj\_out \\
  & \cellcolor[HTML]{909497} up\_blocks1resnets1conv\_shortcut \\
  & \cellcolor[HTML]{909497} up\_blocks2resnets2conv\_shortcut \\
  & \cellcolor[HTML]{909497} down\_blocks2resnets0conv1 \\
  & \cellcolor[HTML]{909497} up\_blocks1attentions1proj\_out \\
  & \cellcolor[HTML]{909497} up\_blocks1attentions1proj\_in \\
  & \cellcolor[HTML]{909497} up\_blocks2resnets1conv\_shortcut \\
  & \cellcolor[HTML]{909497} up\_blocks2resnets0conv\_shortcut \\
  & \cellcolor[HTML]{909497} up\_blocks1attentions0proj\_in \\
  \bottomrule
\end{tabular}}
\end{table}

\begin{table}[h]
\caption{Truth table of the proposed trigger function. When $\mathbf{M}^{S}$ is set as 0, the output sign is based on the input. Otherwise, the sign is determined using $\mathbf{M}^{I}$. }
\label{tab:tbtf}
\setlength\tabcolsep{4pt}
\resizebox{.48\textwidth}{!}{
\begin{tabular}{c|ccc|c}
\toprule
Input Sign ($\mathbf{\tilde{F}}_i$) & $\mathbf{M}^{S}$ & $\mathbf{M}^{I}$ &  $\lvert \mathbf{F}_i \rvert$ & $\mathbf{M}^{S}_i \circ  \mathbf{M}^{I}_i \circ \lvert \mathbf{F}_i \rvert + (1- \mathbf{M}^{S}_i) \circ \mathbf{F}_i$ \\ \midrule
+          & 1                & +    & +           & +           \\
+          & 1                & -    & +           & -           \\
-          & 1                & +    & +           & +           \\
-          & 1                & -    & +           & -           \\ \midrule
+          & 0                & +    & +           & +           \\
+          & 0                & -    & +           & +           \\
-          & 0                & +    & +           & -           \\
-          & 0                & -    & +           & -          \\ \bottomrule
\end{tabular}
}
\end{table}

\subsection{More details about $\mathbf{C}$ in learning objective $\mathcal{L}_{b}$}
\label{sec:constant}
Here, we elucidate the backdoor embedding loss $\mathcal{L}_b$:
\begin{align}
 \mathcal{L}_{b} =   \mathcal{S}_{\text{cos}}( \mathbf{M}^{S}_{j} \circ \mathbf{g}(\mathbf{\tilde{F}}_{i}|i\rightarrow j), \mathcal{R}(v_j) \circ \mathbf{C}).
\end{align}
The learning objective $\mathcal{L}_{b}$ aligns the sign of prediction $ \mathbf{g}(\mathbf{\tilde{F}}_{i}|i\rightarrow j)$ with $\mathcal{R}(v_j) = \mathbf{M}^{S}_j \circ \mathbf{M}^{I}_j $. To ensure the optimal value of $\mathcal{L}_b$ can reach -1, $\mathbf{M}^{S}_j \circ \mathbf{M}^{I}_j$ is normalized by a constant term $\mathbf{C}$,  
\begin{align}
    \mathbf{C} = \frac{|\mathbf{g}^{-}(\mathbf{\tilde{F}}_i|i\rightarrow j)|}{||\mathbf{g}^{-}(\mathbf{\tilde{F}}_i||i\rightarrow j)||},
\end{align}
where $\mathbf{g}^{-}(\cdot|i \rightarrow j)$ refers to $\mathbf{g}$ with stop gradient operator. To better understand the impact of $\mathbf{C}$, we explain it with a toy case:
$\mathbf{g}(\mathbf{\tilde{F}}_i|i \rightarrow j) = \begin{pmatrix} 
-3  \\
 4  \\
-5  \\
 2  \\
\end{pmatrix} $, $
    \mathbf{M}^{S}_j = \begin{pmatrix}
 1  \\
 0  \\
 1  \\
 1  \\
\end{pmatrix}$, $
    \mathbf{M}^{I}_j = \begin{pmatrix}
 -1  \\
 1  \\
 -1  \\
 1  \\
\end{pmatrix}$. 
Then, if the constant term is removed, $\mathcal{L}_{b}$ is calculated by:

\begin{equation}
\begin{aligned}
    \mathbf{M}^{S}_j \circ \mathbf{g}(\mathbf{\tilde{F}}_i|i \rightarrow j) &= \begin{pmatrix}
 -3  \\
 0 \\
 -5  \\
 2  \\
\end{pmatrix}, \\ 
\mathbf{M}^{S}_j \circ \mathbf{M}^{I}_j &= \begin{pmatrix}
 -1  \\
 0 \\
 -1  \\
 1  \\
\end{pmatrix}, \\ 
-\mathbf{S}_{\text{cos}} (\mathbf{M}^{S}_j \circ \mathbf{g}(\mathbf{\tilde{F}}_i|i\rightarrow j),\mathbf{M}^{S}_j \circ \mathbf{M}^{I}_j) &= \\
-\frac{10}{6.164 \times 1.732}& = -0.937. \\
\end{aligned}
\end{equation}
In this case, although $\mathbf{M}^{S}_j \circ \mathbf{g}(\mathbf{\tilde{F}}_i|i \rightarrow j)$ has the same sign as $\mathbf{M}^{S}_j \circ \mathbf{M}^{I}_j$, the corresponding loss value remains larger than $-1$, providing an inaccurate values for learning. In other words, different instances of $\mathbf{g}(\mathbf{\tilde{F}}_i|i \rightarrow j )$ can yield the correct signs but still produce $\mathcal{L}_{b}$ with varying values. 

We then examine the same case but with constant terms as below: 
\begin{equation}
\begin{aligned}
   \mathbf{C} =  \frac{|\mathbf{g}^{-}(\mathbf{\tilde{F}}_i|i\rightarrow j)|}{||\mathbf{g}^{-}(\mathbf{\tilde{F}}_i|i \rightarrow j)||} &= \begin{pmatrix}
 0.487  \\
 0.811  \\
0.324  \\
\end{pmatrix}, \\ 
-\mathbf{S}_{\text{cos}} (\mathbf{M}^{S}_j \circ \mathbf{g}(\mathbf{\tilde{F}}_i|i \rightarrow j),\mathbf{M}^{S}_j \circ \mathbf{M}^{I}_j  \circ   \mathbf{C}) &=  \\
-\frac{6.164}{6.164 \times 1} &= -1. \\
\end{aligned}
\end{equation}
Considering the constant term, the optimal results of $\mathcal{L}_{b}$ is fixed to $-1$, indicating correct alignment between the loss function and our goal. 

\subsection{Theoretical Analysis}
\label{sec:proof}

\section{Theoretical Analysis}
\label{sec:theoretical}
Our pilot study reveals that the busy layer causes performance deterioration after fine-tuning. Therefore, we used the metric $\mathcal{C}$---the number of busy layers encountered during verification---to analyze the stability of the triggers. A robust approach achieves verification with reduced $\mathcal{C}$.  We theoretically prove that our method can reduce the number of visits to busy layers by four times:

\noindent
\textbf{Theorem. 1} \textit{If triggers and responses are embedded in the model's input and output, respectively, $\mathbb{E}[\mathcal{C}] = \mathcal{C}_b$. }

\begin{figure}[tbp]
    \centering
    \includegraphics[width=0.5\textwidth]{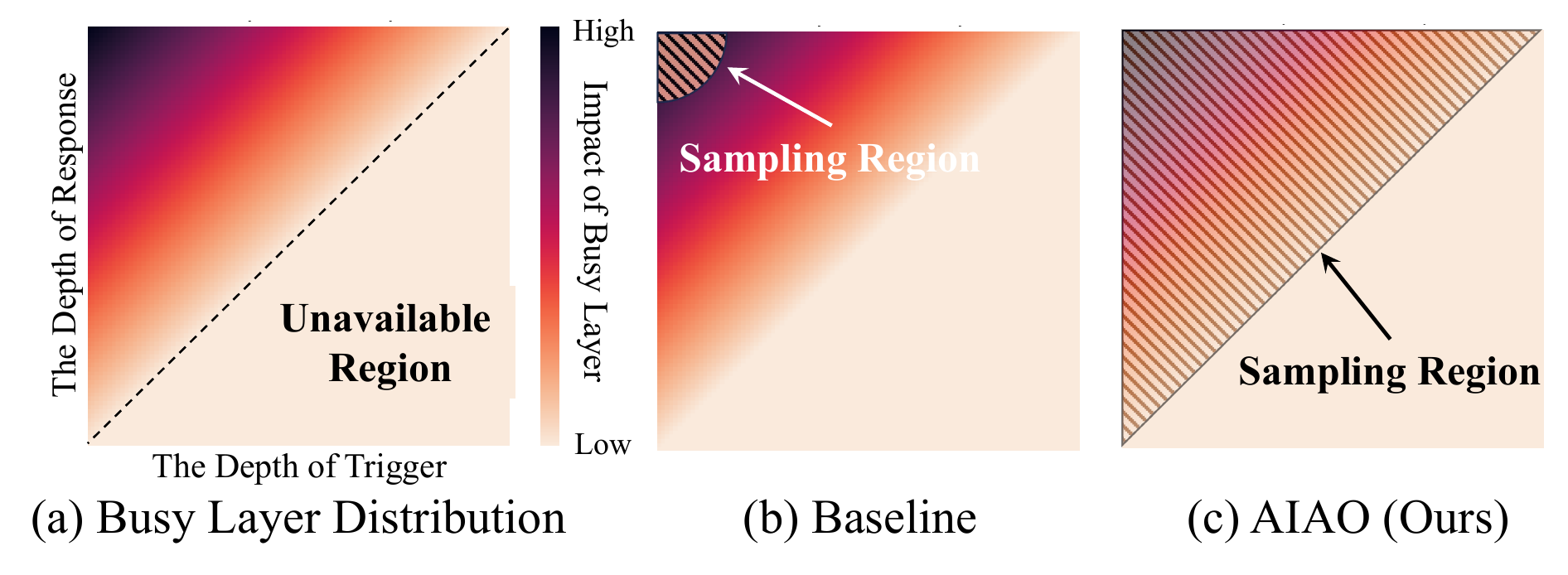}
   \caption{\textbf{Intuitive Explanation of the Proposed Method.} Original methods introduce triggers to the input and detect specific responses in the output. However, after fine-tuning, this verification process encounters all busy layers and thereby suffers from their dramatic changes. Suppose that a busy layer follows a uniform distribution across the depth of the model. The sampling region of this baseline will be limited to the top left corner, where the busy layer distribution achieves the highest density. 
   In contrast, our proposed method allows us to apply triggers at any position and detection can occur at any subsequent positions behind the trigger. By sampling various sub-paths and averaging their verification results, the proposed method can suppress the influence of busy layers. 
   } 
   \label{fig:sample_region}
\end{figure}

\noindent
\textbf{Theorem. 2} \textit{Under the Arbitrary-In-Arbitrary-Out strategy (AIAO), the trigger and response positions follow the conditional uniform distributions. Here, $\mathbb{E}[\mathcal{C}] = \frac{1}{4}\mathcal{C}_b$.}

$\mathcal{C}_b$ is a constant number, presenting the total busy layer numbers within the full DM.

\noindent
\textbf{Intuition}
This theoretical analysis can be explained from the viewpoint of sampling areas. As shown in Fig. \ref{fig:sample_region} (a), when the busy layer is assumed to be uniformly distributed along the model, the probability density of the busy layer is only related to the length between the in-layer and the out-layer. Therefore, the density can be presented as a triangular region. Considering that in traditional approaches, the trigger (in-layer) and the response (out-layer) are located farthest from each other, their sampling area is represented by a point in the upper left corner (see Fig. \ref{fig:sample_region} (b)). In contrast, our sampling area covers the entire feasible region. This indicates that the expectation of our approach should be lower than the baseline in terms of the impact of the busy layer.

\noindent
\textbf{Preliminary}
A watermarked generative model $\mathcal{G}(\cdot)$ consists of $n$ layers where the $i$-th layer refers to $v_i$. After fine-tuning, $\mathcal{C}_b$ layers change, which are grouped into busy layers. The existing observations show that the busy layers are unevenly distributed throughout the depth of the model. As a reasonable assumption, we model the distribution of the busy layers as a uniform distribution $\mathbb{U}(1,\mathbf{L})$, where each busy layer is independently and identically distributed (i.i.d.). The expectation $\mathcal{\mathbb{E}}(\mathcal{C}|v_i,v_j,\mathcal{C}_b)$ distributed within the layers $\{ v_i, v_{i+1},...,v_{j-1},v_{j}\} $ can be calculated by 
\begin{align}
    \mathbb{E}(\mathcal{C}|v_i,v_j,\mathcal{C}_b) = \mathcal{C}_b \frac{j-i}{\mathbf{L}} = \mathcal{C}_b \frac{j-i}{n-1},
\end{align}
where $i$ and $j$ represent random variables that denote the in-layer and out-layer, respectively.

\noindent
\textit{Proof of Theorem. 1:} Since the in-layer and out-layer are fixed, their probability density function (PDF) is given as 
\begin{equation}
    \begin{aligned}
        f_{\text{in}}(v) = 
        \begin{cases}
            1 & \text{if } v = v_1\\
            0 & \text{otherwise}
        \end{cases},
    \end{aligned}
\end{equation}
\begin{equation}
    \begin{aligned}
        f_{\text{out}}(v) = 
        \begin{cases}
            1 & \text{if } v = v_{n}\\
            0 & \text{otherwise}
        \end{cases},
    \end{aligned}
\end{equation} 
where $ f_{\text{in}}(v)$ and $ f_{\text{out}}(v)$ are independent; $v_1$ and  $v_{n}$ are the first layer and last layer in $\mathcal{G}$, respectively. The expectation of the number of busy layers $\mathbb{E}(\mathcal{C}|\mathcal{C}_b,f_s(\cdot),f_e(\cdot))$ can be then computed by 
\begin{equation}
\begin{aligned}
   \mathbb{E}(\mathcal{C}|\mathcal{C}_b,f_{\text{in}},f_{\text{out}}) &=  \int\int f_{\text{in}}(v)f_{\text{out}}(v)\mathbb{E}(\mathcal{C}|v_i,v_j,\mathcal{C}_b) \, dv_i dv_j \\
   &=  \mathcal{C}_b.
\end{aligned}
\end{equation}
Note that even with multiple sampling, as in the case of AIAO, the expectation of the baseline still converges to $\mathcal{C}_b$, which cannot be further reduced. 

\noindent
\textit{Proof of Theorem. 2:}
In the AIAO strategy, the in-layer PDF is 
\begin{equation}
    \begin{aligned}
        \Bar{f}_{\text{in}}(v) = \frac{1}{n-1} \quad s.t. \, v\in \mathcal{G}.
    \end{aligned}
\end{equation}
Considering that the response should be placed behind the trigger, the out-layer follows a conditional PDF $\Bar{f}_{\text{out}}(v|v_i)$,
\begin{equation}
    \begin{aligned}
        \Bar{f}_\text{out}(v|v_i) = \frac{1}{n-i} \quad s.t. \, v \in \mathbf{g} (\cdot|i\rightarrow n).
    \end{aligned}
\end{equation}
Based on Bayes' theorem, $\mathbb{E}(\mathcal{C}|\mathcal{C}_b,\Bar{f}_{\text{in}}(\cdot),\Bar{f}_{\text{out}}(\cdot))$ is obtained as 
\begin{equation}
    \begin{aligned}
        \mathbb{E}(\mathcal{C}|\mathcal{C}_b,&\Bar{f}_{\text{in}}(\cdot),\Bar{f}_\text{out}(\cdot)) \\
        &= \int \Bar{f}_{\text{out}}(v|v_i) \Bar{f}_\text{in}(v) \mathbb{E}(\mathcal{C}|v_i,v_j,\mathcal{C}_b) \,dv_i dv_j \\
        &= \frac{\mathcal{C}_b}{(n-1)^2} \int_{1}^{n} \int_{i}^{n} \frac{j-i}{n-i} \,di\,dj  \\
        &= \frac{\mathcal{C}_b}{(n-1)^2} \int_{1}^{n} \frac{1}{n-i}(\frac{1}{2}n^2 - ni + \frac{1}{2}i^2)di \\
        &= \frac{\mathcal{C}_b}{(n-1)^2} \int_{1}^{n} \frac{n-i}{2} di \\
        &= \frac{\mathcal{C}_b}{(n-1)^2} \frac{1}{4}(n-1)^2 = \frac{1}{4}\mathcal{C}_b.
    \end{aligned}
\end{equation}
Based on Theorem. 1, $\mathbb{E}(\mathcal{C})$ of the baseline equals to $\mathcal{C}_b$, which is four times higher than that of our methods. Since a smaller  $\mathbb{E}(\mathcal{C})$ implies greater stability, we theoretically demonstrate that the AIAO strategy can enhance resilience to the forgetting caused by fine-tuning. 

\begin{figure}[tbp]
    \centering
    \includegraphics[width=0.48\textwidth]{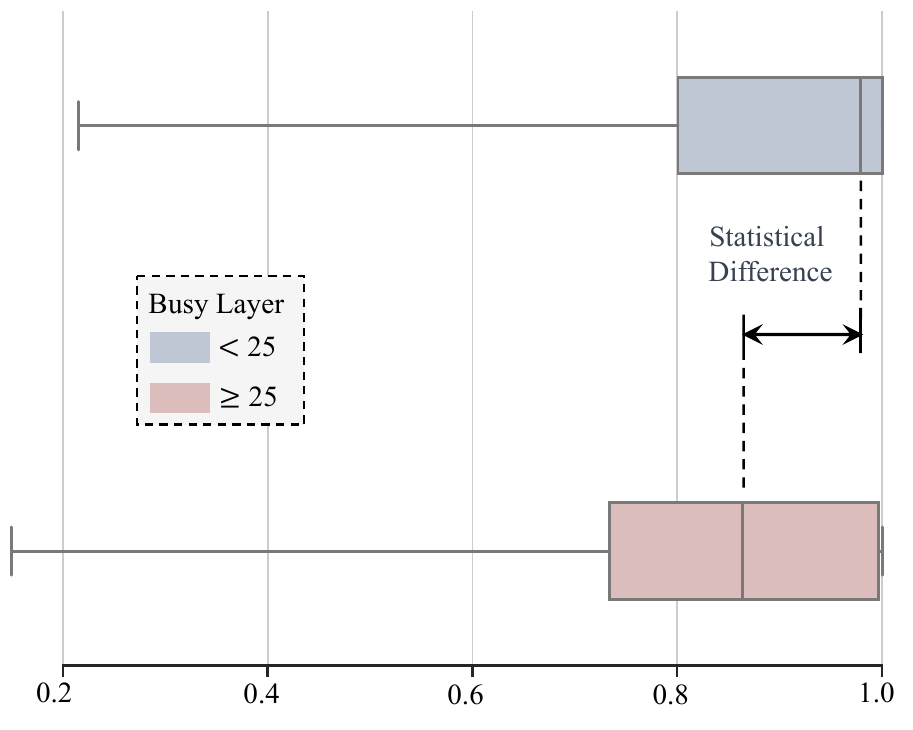}
   \caption{
   \textbf{Empirical Validation of the Negative Impacts of Busy Layers.} Our method samples multiple in-layers and out-layers to construct sub-paths. We test the distribution response success rates (after fine-tuning in clean data) for different sub-paths. The top 50 layers exhibiting the most significant changes during fine-tuning are designated as the busy layers.  Sub-paths are categorized into two groups based on whether they encounter less or more than 25 busy layers. These experiments are based on 25k input samples.  Verification with fewer busy layers consistently yields higher accuracy.
   The experiments are carried out on the unconditional image generation task (dog$\rightarrow$cat)
   } 
   \label{fig:app_busy}
\end{figure}

\noindent
\textbf{Empirical study on busy layers}
Our theoretical analysis is based on the negative impact of busy layers. 
We empirically validate the supposition that the busy layer causes the forgetting of the watermarks. We identify the busy layers as the 50 layers with the most significant shifts after fine-tuning. By sampling 25k trigger-response pairs, we analyze the response success rates in the presence of varying numbers of busy layers. In Fig. \ref{fig:app_busy}, we present the distribution of response success rates for cases with more than or less than 25 busy layers. The results show that responses are more precise when there are fewer busy layers in agreement with our observation that busy layers are the key to forgetting triggers.

\label{sec:verify}
\begin{table}[]
\caption{Summarization of the Competing Methods Discussed in the Paper.}
\label{tab:summ_compare}
\Huge
\resizebox{.49\textwidth}{!}{
\begin{tabular}{cccc}
\toprule
Protected Model                    & Category                               & Method           & Publication \\ \midrule
\multirow{5}{*}{Fine-tune DM} & \multirow{3}{*}{Backdoor Watermarking}   & WDP\cite{peng2023protecting}              & Arxiv'23    \\ \cmidrule(lr){3-4} 
                                   &                                        & WatermarkDM \cite{zhao2023recipe}     & Arxiv'23    \\ \cmidrule(lr){3-4} 
                                   &                                        & FixedWM \cite{liu2023watermarking}         & Arxiv'23    \\ \cmidrule(lr){2-4} 
                                   & Image Watermarking                     & Stable Signature \cite{fernandez2023stable} & ICCV'23     \\ \midrule
\multirow{5}{*}{w/o Fine-tuning DM}     & Distributional Distance                & FID \cite{heusel2017gans}             & NIPS'17     \\ \cmidrule(lr){2-4} 
                                   & \multirow{2}{*}{Image Similarity}      & DINO-v1 \cite{caron2021emerging}         & ICCV'21     \\ \cmidrule(lr){3-4} 
                                   &                                        & MoCo-v3 \cite{fan2021multiscale}         & ICCV'21     \\ \cmidrule(lr){2-4} 
                                   & Model Attribution                      & GAN-Guards \cite{hu2023ownership}      & Arxiv'23    \\ \bottomrule
\end{tabular}
}
\end{table}

\subsection{Experimental Setup on Text-to-Image Generation}
\label{sec:at2i}

We use SD-1.4 \cite{rombach2022high} as the source model. The source model learns to embed identifiers on the MS-COCO dataset (2017) \cite{lin2014microsoft}. The MS-COCO dataset comprises 118K pairs of images and captions for training along with an additional 5K pairs for validation. 
To evaluate the stability of the compared methods over fine-tuning, we evenly partition the MS-COCO train set into two subgroups, namely, COCO-A and COCO-B. COCO-A is dedicated to embedding triggers, while COCO-B serves as the new data for testing the source models over in-distribution fine-tuning. For the out-of-distribution fine-tuning, we use CUB \cite{WahCUB_200_2011} as the target data, which contains 5994 image-caption pairs in the training set. 

\noindent
\textbf{Training Protocol.}
We adopt AdamW optimizer \cite{loshchilov2017decoupled} and a noise scheduler of DDPM \cite{ho2020denoising} with an initial learning rate of $1 \times 10^{-5}$ during training. The batch size is 128, and the maximum number of training steps is 3k for embedding our backdoor and 6k for fine-tuning-based removal.

\begin{table*}[]
\setlength\tabcolsep{12pt}
\caption{Unconditional generation performance of the DM \cite{rombach2022high} on AFHQ and LSUN datasets, where the DM is equipped with various ownership protection methods. The generation performance is measured by FID $\downarrow$.  The baseline refers to SD-1.4 \cite{rombach2022high}  without backdoor embedding. }
\label{tab:uncondition_generation}
\resizebox{.99\textwidth}{!}{
\begin{tabular}{c|l|c|cccc}
\toprule
\multirow{2}{*}{Protocol}  &\multirow{2}{*}{Ownership protection  method}                                                  & \multirow{2}{*}{Source Model}                   & \multicolumn{4}{c}{ Fine-tuning on Downstream Dataset} \\ \cmidrule(lr){4-7}
                                      &                         &      & 0.5k steps    & 1k steps      & 1.5k steps      & After Fine-tuning                 \\ \midrule
\multirow{3}{*}{Cat $\rightarrow$ Dog} & Baseline         &   15.84              &                   28.33&                   31.75&                   30.38&                    16.28\\ 
                                        & WDP \cite{peng2023protecting}        &  16.05               &                   24.76&                   25.29&                   21.63&                   16.16\\ 
                                                                                &\cellcolor[HTML]{EFEFEF}AIAO                        &                 \cellcolor[HTML]{EFEFEF}15.29&                   \cellcolor[HTML]{EFEFEF}33.51&                   \cellcolor[HTML]{EFEFEF}24.07&                   \cellcolor[HTML]{EFEFEF}23.14& \cellcolor[HTML]{EFEFEF}14.93\\ \midrule
\multirow{3}{*}{Dog $\rightarrow$ Cat} & Baseline         &                 19.65&                   11.94&                   15.06&                   14.03&                   10.17\\ 
                                        & WDP \cite{peng2023protecting}                        &                 30.47&                   14.98&                   17.09&                   10.73&                   10.73\\ 
                                                                                & \cellcolor[HTML]{EFEFEF}AIAO                        &                 \cellcolor[HTML]{EFEFEF}25.40&                   \cellcolor[HTML]{EFEFEF}10.00&                   \cellcolor[HTML]{EFEFEF}10.33&                   \cellcolor[HTML]{EFEFEF}8.71& \cellcolor[HTML]{EFEFEF}7.36\\ \bottomrule
                                                                                
\end{tabular}
}
\end{table*}

\begin{table*}[]
\caption{Text-to-image generation performance of the watermarked DMs on Caltech-UCSD Birds (CUB) dataset with In-Distribution Fine-tuning Protocol, where \textcolor{fid}{FID}$\downarrow$ and \textcolor{clip}{CLIP}$\uparrow$ scores measure the generation performance. The baseline refers to SD-1.4 \cite{rombach2022high}  without backdoor embedding. }
\label{tab:cub_generation}
\setlength\tabcolsep{15pt}
\resizebox{.99\textwidth}{!}{
\begin{tabular}{l|cccc}
\toprule
\multirow{2}{*}{Method}              & \multicolumn{4}{c}{Finetuning-based Removal} \\ \cmidrule(lr){2-5}
                                     & 0.5k steps& 1k steps& 1.5k steps& 2k steps\\ \midrule
Baseline   &  \textcolor{fid}{13.25}/\textcolor{clip}{26.23}&   \textcolor{fid}{12.31}/\textcolor{clip}{25.83}&       \textcolor{fid}{10.94}/\textcolor{clip}{25.88}&     \textcolor{fid}{10.77}/\textcolor{clip}{25.94}\\  \midrule

WatermarkDM (caption-watermark)  \cite{zhao2023recipe}      &  \textcolor{fid}{13.10}/\textcolor{clip}{26.07}&   \textcolor{fid}{12.09}/\textcolor{clip}{25.88}&       \textcolor{fid}{10.90}/\textcolor{clip}{26.03}&     \textcolor{fid}{10.86}/\textcolor{clip}{25.94}\\ 
FixedWM (caption-watermark) \cite{liu2023watermarking}                                &        \textcolor{fid}{13.81}/\textcolor{clip}{26.07}&  \textcolor{fid}{11.46}/\textcolor{clip}{25.96}&      \textcolor{fid}{10.96}/\textcolor{clip}{25.95}&    \textcolor{fid}{11.01}/\textcolor{clip}{26.05}\\ 
\cellcolor[HTML]{EFEFEF}AIAO (Ours)  &  \cellcolor[HTML]{EFEFEF}\textcolor{fid}{13.70}/\textcolor{clip}{25.94}&  \cellcolor[HTML]{EFEFEF}\textcolor{fid}{11.91}/\textcolor{clip}{25.95}&   \cellcolor[HTML]{EFEFEF}\textcolor{fid}{10.64}/\textcolor{clip}{25.67}&  \cellcolor[HTML]{EFEFEF}\textcolor{fid}{10.80}/\textcolor{clip}{25.94}\\ 
\bottomrule
\end{tabular}
 }
\end{table*}

\noindent
\textbf{Baselines}
We compare our method with caption-watermark approaches: WatermarkDM \cite{zhao2023recipe} and FixedWM \cite{liu2023watermarking}. WatermarkDM \cite{zhao2023recipe} takes a special fixed caption as the trigger input, \cite{zhao2023recipe}, and  FixedWM \cite{liu2023watermarking} takes a caption containing a special word \cite{zhai2023text,liu2023watermarking}. %
In addition, we examine Stable Signature \cite{fernandez2023stable}, which directly embeds watermarks into the generated samples. All the baselines used are summarized in Table \ref{tab:summ_compare}. 

\subsection{Experimental Setup on Unconditional Image Generation}
\label{sec:uig}
We use AFHQ-Dog \cite{choi2020stargan}, AFHQ-Cat \cite{choi2020stargan}, LSUN-Church \cite{yu2015lsun}, and LSUN-Bedroom \cite{yu2015lsun} datasets. Each subdataset of AFHQ contains 5k animal images, LSUN-Church contains 12.6k, and LSUN-Bedroom consists of 3,033k images. The diffusion model is initially trained on one of these datasets to learn the watermarks. Subsequently, another dataset is used to fine-tune this model for evaluating the robustness of the triggers. To ensure the DM where the watermark is inserted has satisfactory generative performance, we initialize the weights from the pre-trained SD-1.4 \cite{rombach2022high} with a fixed caption input,  i.e., the text encoder was banned in this case. 
The training protocol is the same as that used for the text-to-image generation except that the maximum training step is set to 6k. 

\noindent
\textbf{Baselines.}
For unconditional DM, we compare two types of ownership protection methods. In particular, since WatermarkDM \cite{zhao2023recipe} and FixedWM \cite{liu2023watermarking} are not designed for unconditional DM, we compare our method with a noise-watermark ownership protection method \ie, WDP \cite{peng2023protecting}. WDP generates an image marked by a fixed word when a trigger noise is fed as an input. 
In addition, as shown in Table.\ref{tab:summ_compare}, a few image similarity algorithms may be treated as ownership protection methods since they can be used to measure the image similarity between the test model and the protected model's generated images. Hence, we use MoCo-V3 \cite{fan2021multiscale} and DINO-V1 \cite{caron2021emerging} to measure the similarity in latent space. Meanwhile, we use the FID \cite{heusel2017gans} to verify the ownership by estimating the distance between the generated distribution of the protected and test model. Some studies \cite{huang2023can,hu2023ownership}, motivated by model attribution \cite{mirsky2021creation}, learn the pattern hidden in the generated images to identify the ownership. We use GAN-Guards \cite{hu2023ownership} as the representative approach for comparison.

\subsection{More Image Generation Comparison}
\label{sec:generative_perform_uncond}

We evaluate the impact of our methods on text-to-image DM \cite{rombach2022high}. In particular, we separately incorporate the DM  with  WatermarkDM, FixedWM, and our method via fine-tuning on COCO-A. All incorporated DMs are subjected to fine-tuning-based removal using an out-of-distribution dataset (CUB) to challenge these ownership protection methods. As shown in Table \ref{tab:cub_generation},  the generation performance of our method is comparable with that of the baseline (\ie, the original DM), even under the challenging fine-tuning-based removal settings with out-of-distribution data.

 We evaluate the influence of our method on the generation performance of DMs.  As shown in Table \ref{tab:uncondition_generation}, the unconditional DM incorporated with our method achieves better image quality than that incorporated with WDP, indicating that our method has a less negative influence on DM.

\subsection{Implementation Details of WDP \cite{peng2023protecting}}
\label{sec:app_wdp}
\begin{figure}[tbp]
    \centering
    \includegraphics[width=0.48\textwidth]{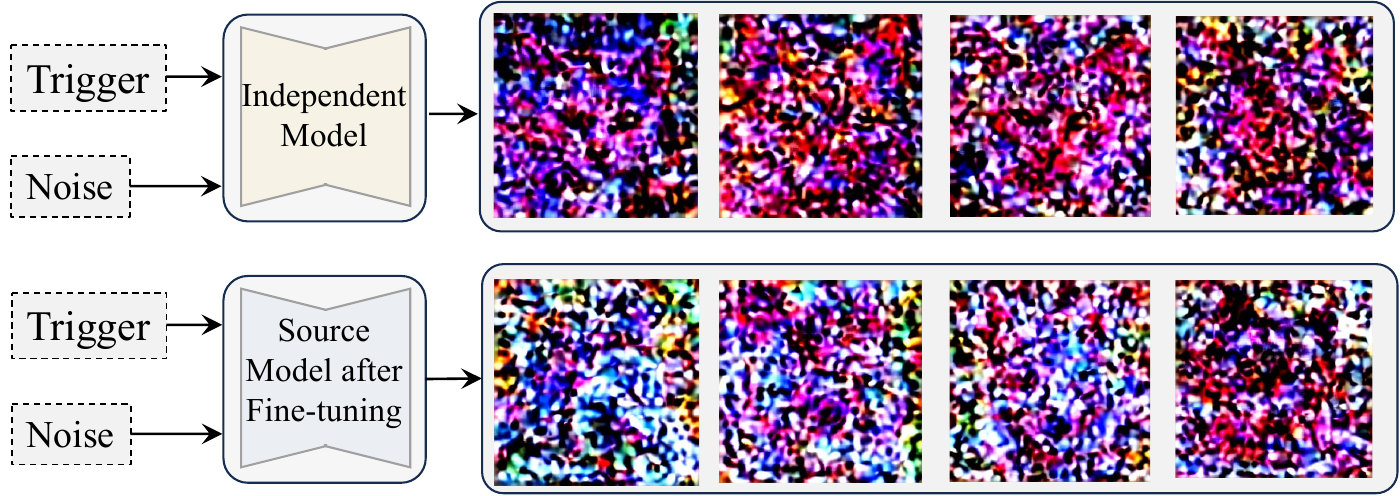}
   \caption{\textbf{Illustration of the failure for WDP.} In (a), given the trigger as input, the independent model cannot output any meaningful images.  In (b), following fine-tuning-based removal, the model cannot generate watermarked images but still shows some statistical difference over (a), which can reach $\sim$80\% verification success rates. However, this verification is not encouraged since the generated images are meaningless.  
   } 
   \label{fig:visual_wdp}
\end{figure}

Our experiment shows that even with a low response success rate, WDP maintains moderate verification performance. To explain this, we visualize images generated by independent and protected models when the trigger is activated, as shown in Fig. \ref{fig:visual_wdp}. Although the protected model fails to generate meaningful images after fine-tuning, there are some visible differences between the images generated by different models.

\subsection{More about Backdoor Attacks}
Most existing methods, which are applied for discriminative models, design triggers for inputs (\eg, images) and define abnormal behavior for the model outputs. 
However, some approaches \cite{tang2020embarrassingly,li2021deeppayload} use an additional neural network to learn specified behaviors when direct access to the training process of the model is unavailable. 
Techniques for more stealthy operations \cite{dumford2020backdooring,li2021deeppayload} embed triggers within weight perturbations. Despite these advancements, designing triggers and activating backdoors in the feature space (intermediate layer) remains under-explored. Salem et al. \cite{salem2020don} propose using the dropout as a trigger, but this approach is incompatible with diffusion processes. 

\end{document}